%% file: acl_latex.tex
\pdfoutput=1

\documentclass[11pt]{article}

\usepackage[]{acl}

\usepackage{times}
\usepackage{latexsym}
\usepackage{adjustbox}
\usepackage[T1]{fontenc}

\usepackage[utf8]{inputenc}

\usepackage[edges]{forest}
\definecolor{lightcoral}{rgb}{0.94, 0.5, 0.5}
\definecolor{lightgreen}{rgb}{0.56, 0.93, 0.56}
\definecolor{harvestgold}{rgb}{0.98, 0.85, 0.40}
\definecolor{brightlavender}{rgb}{0.75, 0.58, 0.89}
\definecolor{capri}{rgb}{0.0, 0.75, 1.0}
\definecolor{carminepink}{rgb}{0.92, 0.3, 0.26}
\definecolor{celadon}{rgb}{0.67, 0.88, 0.69}
\definecolor{darkpastelgreen}{rgb}{0.01, 0.75, 0.24}
\definecolor{purple}{RGB}{144,153,196}
\definecolor{hidden-draw}{RGB}{205, 44, 36}
\definecolor{hidden-blue}{RGB}{194,232,247}
\definecolor{hidden-orange}{RGB}{243,202,120}
\definecolor{hidden-yellow}{RGB}{242,244,193}
\definecolor{hidden-black}{RGB}{20,68,106}
\definecolor{tree-level-1}{RGB}{245,20,85}
\definecolor{tree-level-2}{RGB}{246,86,118}
\definecolor{tree-level-3}{RGB}{248,177,193}
\definecolor{tree-leaf}{RGB}{176,230,198}

\definecolor{Self}{RGB}{255,0,128}
\definecolor{Ensemble}{RGB}{0,127,255}
\definecolor{Iterative}{RGB}{153,51,255}

\definecolor{exemplar1}{RGB}{136,98,148}
\definecolor{exemplar2}{RGB}{148,210,242}
\definecolor{knowledge1}{RGB}{249,219,152}
\definecolor{knowledge2}{RGB}{255,245,220}

\usepackage{nicematrix}
\usepackage{makecell}
\usepackage{xspace,mfirstuc,tabulary}
\usepackage{booktabs}
\usepackage{graphicx}
\usepackage{amssymb}
\usepackage{bbding}
\usepackage{xcolor,colortbl}
\usepackage{multirow}
\usepackage{inconsolata}
\usepackage{enumitem}

\usepackage{pifont}

\newrobustcmd{\B}{\bfseries}
\newcommand*\colourcheck[1]{%
  \expandafter\newcommand\csname #1check\endcsname{\textcolor{#1}{\ding{52}}}%
}
\newcommand*\colourcross[1]{%
  \expandafter\newcommand\csname #1cross\endcsname{\textcolor{#1}{\ding{55}}}%
}

\colourcheck{green}
\colourcross{red}

\newcommand{\eg}{\emph{e.g.}}

\usepackage{microtype}

\usepackage{inconsolata}

\usepackage{graphicx}

\title{The Oscars of AI Theater: A Survey on Role-Playing with Language Models}

\author{
 Nuo Chen$^\spadesuit$
\quad
Yan Wang$^\clubsuit$
\quad 
{\bf Yang Deng$^{\diamondsuit}$}  
{\bf \quad Jia Li$^\spadesuit$}\\
\\
  $^\spadesuit$Hong Kong University of Science and Technology (Guangzhou)\\ Hong Kong University of Science and Technology\\
  $^\clubsuit$ Tencent\\
  $^{\diamondsuit}$Singapore Management University\\
    \texttt{nchen022@connect.ust.hk}, 
    \texttt{yanwang.branden@gmail.com},
    \texttt{jialee@ust.hk}\\}

\begin{document}
\maketitle
\begin{abstract}


This survey explores the burgeoning field of role-playing with language models, focusing on their development from early persona-based models to advanced character-driven simulations facilitated by Large Language Models (LLMs). Initially confined to simple persona consistency due to limited model capabilities, role-playing tasks have now expanded to embrace complex character portrayals involving character consistency, behavioral alignment, and overall attractiveness. 
We provide a comprehensive taxonomy of the critical components in designing these systems, including data, models and alignment, agent architecture and evaluation. This survey not only outlines the current methodologies and challenges, such as managing dynamic personal profiles and achieving high-level persona consistency but also suggests avenues for future research in improving the depth and realism of role-playing applications. The goal is to guide future research by offering a structured overview of current methodologies and identifying potential areas for improvement. Related resources and papers are available at \url{https://github.com/nuochenpku/Awesome-Role-Play-Papers}.

\end{abstract}

\input{sections/1Introduction}

\input{sections/2Background}

\input{figures/catagorization}
\input{sections/4Benchmarks}


\input{sections/6Alignment}
\input{sections/6System_Architecture.tex}
\input{figures/evaluation}
\input{sections/5Evaluation}
\input{sections/6Challengs_Future}
\input{sections/7Conclusion}

\bibliography{custom}




\end{document}

%% file: sections/1Introduction.tex
\section{Introduction}


Today, most large language models (LLMs) \cite{brown2020language, hu2021lora, zeng2022glm, openai2023gpt4,scao2022bloom} are proficient enough to act as assistants, but the ever-expanding desires of humans gradually go beyond this role. A helpful but serious assistant isn't everything in human life. An increasing number of individuals have been instructing LLMs to take on roles they desire, such as movie stars, game characters, or even their own relatives. This practice of aligning LLMs with specific personas or characters is commonly known as \textbf{Role-Playing} \cite{zhang-etal-2018-personalizing,JiangPersonaLLMIT,chen2023large,DBLP:conf/sigir/QianLZGMZLDW21,chen2024personapersonalizationsurveyroleplaying}. If the standard assistant role of LLMs meets the demand for increased productivity, then LLMs for role-playing  aims to fulfill human needs at a psychological and entertainment level. This lively trend underscores the versatility of LLMs and the limitless potential of human imagination in the realm of artificial intelligence.

The requirements for role-playing with language models differ significantly from those of a generic assistant. The primary expectation from a generic assistant is its helpfulness, meaning the LLM should follow the user's instructions and provide the desired responses \cite{serban2016generative,DBLP:conf/sigdial/LowePSP15,DBLP:journals/corr/abs-2201-08239,miller2017parlai,you2022end}. This expectation is also evident in the benchmarks for such models: people often desire an assistant with extensive professional knowledge and strong logical reasoning abilities.

However, when the task comes to role-playing, the most crucial criterion is the LLM's ability to align with specific personas or characters \cite{tu2024charactereval, tu2023characterchat, chen2023large}. In other words, humans expect the LLMs to interact with them in a manner consistent with a specific role. This expectation introduces a fascinating dynamic that can sometimes contradict the traditional notion of helpfulness. For instance, consider a scenario where the role to be played is that of the user's adversary or enemy. In such a case, being helpful becomes a contradictory metric. The more helpful the model is, the less effective it becomes at role-playing.

In the era marked by the advent of sequence to sequence learning \cite{shang2015neural}, researchers began exploring the potential of neural networks to generate dialogue responses that are consistent with both the given context and the portrayed persona \cite{zhang-etal-2018-personalizing,JiangPersonaLLMIT, dinan2018wizard}.  For instance, \citet{zhang-etal-2018-personalizing} utilizes a generative profile memory network to generate personal responses.  This initial motivation laid the groundwork for the following role-playing works. Subsequent advancements with the introduction of models such as BERT \cite{kenton2019bert} brought significant evolutions to the use of language models for role-playing, despite certain inherent limitations. In this period, due to the improved but still developing generative capabilities of these pre-trained language models (PLMs) \cite{liu2019roberta, raffel2020exploring}, role-playing is largely focused on achieving persona consistency through simpler, more straightforward persona roles, often termed as \textbf{persona-based role-playing}. This is partly due to the nature of the personalized information available in datasets at the time, which was often succinct and sparse, as seen in resources like the Persona-Chat dataset \cite{zhang-etal-2018-personalizing}.  PLMs are fine-tuned on these datasets to produce responses that aligned with the limited personal information provided, striving for a balance between consistent understanding and competent dialogue generation.

As the field progressed into the era of LLMs \cite{openai2023gpt4, touvron2023llama, zeng2022glm}, a paradigm shift occurred. The enhanced comprehension and generative abilities of LLMs expand the scope of role-playing tasks far beyond simple persona adherence. Current research in role-playing no longer confines itself to rigid personas. Instead, it delves into more nuanced aspects of role enactment, such as character consistency, behavioral alignment, and overall attractiveness of the character portrayal \cite{chen2023large,shao2023character,tu2024charactereval}, named as \textbf{character-based role-playing}. These dimensions aim to create more immersive and believable character simulations that maintain continuity over interactions and adapt dynamically to dialogue contexts. The progress of LLM-based role-playing leading to a rapid expansion in academic research and the development of practical applications, exemplified by platforms like Character AI\footnote{https://character.ai/}, Xingye\footnote{https://www.xingyeai.com/}, and Maopaoya\footnote{https://maopaoya.com/}.

Despite the promising potential of role-playing with language models, research in this domain remains in its early stages, marked by both complexities and challenges. The goal of this survey is to understand the crucial mechanisms and methodologies that enable role-playing through text-based interactions. To achieve a thorough understanding, we introduce a detailed taxonomy to systematically examine the critical components involved in designing role-playing language models.  The proposed taxonomy includes: \textbf{Data}, \textbf{Models \& Alignment}, \textbf{Agent Architecture}, and \textbf{Evaluation}. This framework aims to not only detail how role-playing functions within these systems but also to highlight how it can be optimized and evaluated for a variety of applications.


In summary, \S \ref{background} first introduces  the preliminary of our survey, like the evolution of language models and key components.
\S \ref{dataset} provides a comparative overview of the current role-playing data resources, highlighting their unique characteristics and applicability in different scenarios.  
For models \& alignment, \S  \ref{alignment} systematically reviews the foundational models of role-playing and
summarizes the past alignment approaches, offering insights into their strengths and weaknesses.  For Agent architecture, \S \ref{architecture} details the important modules that impact the effectiveness and generalization of role-playing language model agents, including memory, planning, action. Then, \S \ref{evaluation} compiles comprehensive evaluation standards and metrics, encompassing both subjective and objective metrics, along with their respective advantages and disadvantages.  Finally, \S \ref{challenge} delves into the 10 main challenges that persist in role-playing and envisages possible solutions that could pave the way for more advanced and nuanced systems. 



%% file: sections/2Background.tex
\section{Background}

\begin{figure*}[!t]
\vspace{-10pt}
\centering
\includegraphics[width=0.95\linewidth]{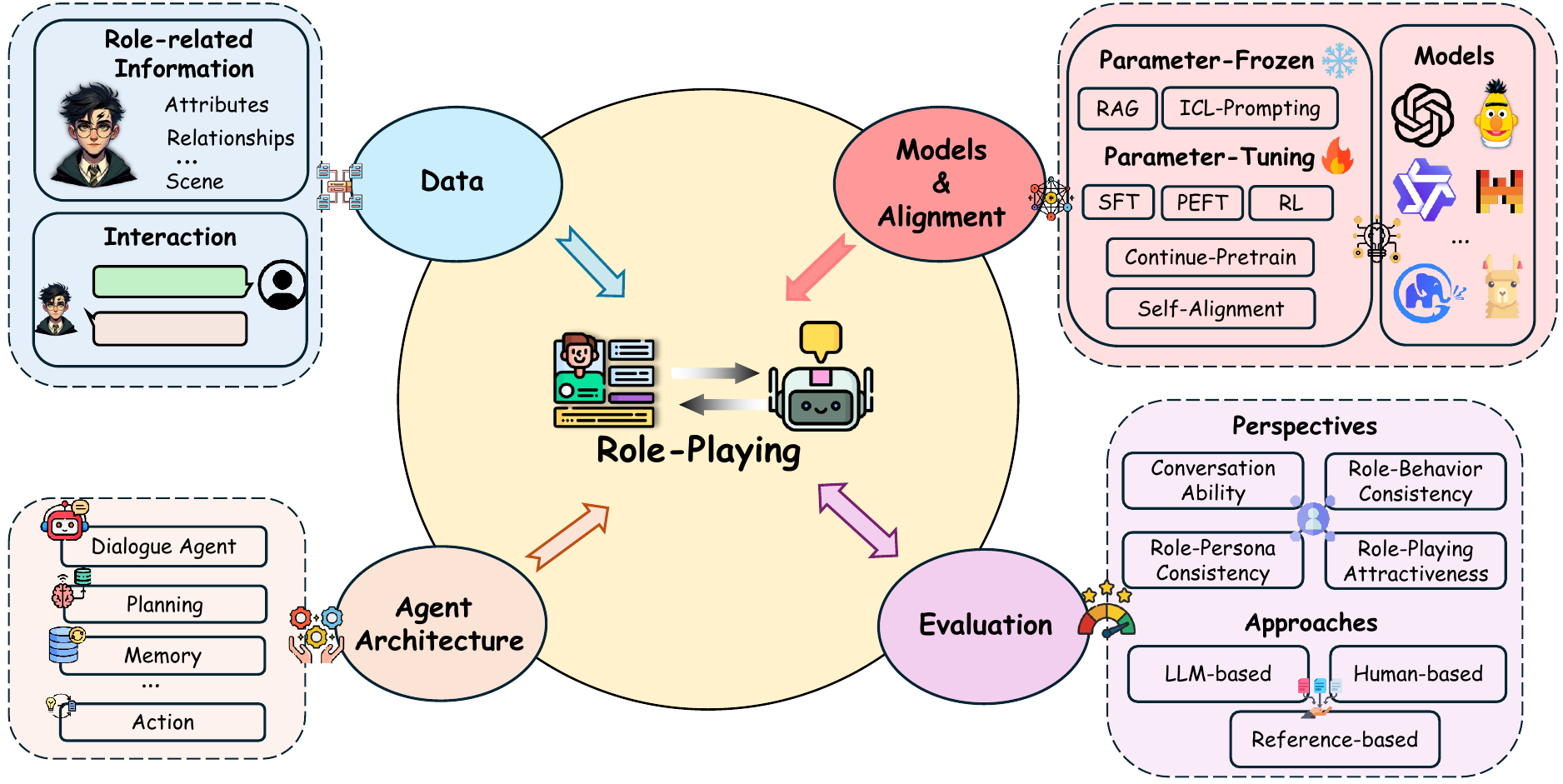}
\caption{Key components in role-playing with language models.
}
\label{fig:case}
\end{figure*}

\label{background}

\subsection{PLMs and LLMs}

Despite the absence of a universally accepted definition for Large Language Models (LLMs), this paper proposes a specific delineation for LLMs as referenced within our analysis. Distinguished by both model scale and training methodologies, our definition builds upon distinctions made by two seminal surveys on LLMs \citep{zhao2023survey,yang2023harnessing}, differentiating LLMs from earlier Pre-trained Language Models (PLMs) based on the magnitude of model parameters and the scope of pre-training data. Specifically, LLMs refer to expansive models with parameters in the billions, pre-trained on extensive datasets, in contrast to PLMs, which are characterized by their relatively moderate parameter sizes in the millions and the capability for efficient task-specific fine-tuning to enhance performance on downstream tasks. 

Notable PLMs include BERT \cite{kenton2019bert}, GPT-2 \cite{Radford2019LanguageMA}, BART \cite{lewis2020bart}, and Roberta \cite{liu2019roberta},
 whereas leading examples of LLMs, such as GPT-3 \citep{brown2020language}, PaLM \citep{chowdhery2022palm}, Mistral \cite{jiang2023mistral}, and LLaMA \citep{touvron2023llama}, primarily utilize decoder-only architectures. The progression in architectural and training innovations for LLMs has facilitated the emergence of advanced capabilities \citep{wei2022emergent}, empowering them to tackle complex problems in few-shot or zero-shot settings through methodologies like in-context learning \citep{radford2021learning,dong2022survey} and chain-of-thought reasoning \citep{wei2022chain}. After reading this review, you will find that, in fact, the capabilities of zero-shot and few-shot learning are the main reasons for the increasing attention and interest in the domain of role-playing over the past two years.

\subsection{Key Components}
    Employing language models for role-playing involves several critical factors that significantly influence their effectiveness and personalization capabilities. In this survey, we analyze several components that are essential in shaping the development, deployment, and continuous improvement of role-playing language models:
\begin{itemize}[leftmargin=*]

    \item \textbf{Data}: The diversity and complexity of the data used in role-playing are foundational, influencing the models' ability to generate authentic and personalized interactions. Role-related information in current datasets ranges from structured texts and simple sentences to rich compilations of detailed personal information like attributes, relationships, and even nuanced understandings of characters across different timelines. The sophistication of a language model in handling role-playing scenarios directly correlates with the complexity of the targeted dataset. Models trained on simpler data might generate broad, generic responses, while those trained on more complex datasets are capable of producing dynamic dialogues that reflect specific character nuances. Hence, the selection and design of the dataset are critical in role-playing, directly impacting effectiveness in delivering engaging and personalized conversational experiences.


    \item \textbf{Models \& Alignment}:
    Undoubtedly, the backbone models play a pivotal role in the role playing's operational efficiency. From traditional neural networks to language models, the choice of model influences the system's understanding, generation, and adaptation capabilities. The recent advent of LLMs has brought significant advancements in this area, achieving high levels of personalization and coherence.
    
    Meanwhile, alignment approaches focus on ensuring that the role-playing models' responses align with the intended role. These methods range from rule-based systems that manually map responses to personas, to dynamic learning mechanisms that adapt responses based on interaction history and persona data. Technically,  we divide them into Parameter-Tuning: Post-training, Supervised Fine-Tuning (SFT), and Reinforcement Learning; and Parameter-Frozen: In-context learning prompting and Retrieval-Augmented Generation (RAG).


    \item \textbf{Agent Architecture}: 
Currently, the development of Role-Playing Language Agents (RPLAs) marks a new evolution. These agents extend the basic framework of role-playing by integrating both interactive and autonomous behaviors, enabling them to not only personify specific characters but also engage proactively in complex and evolving scenarios.
Effective RPLAs require a comprehensive system architecture that includes several key modules: memory, for recalling and utilizing past interactions; planning, for strategic decision-making; and action, which encompasses both generating role-related responses and using tools. Such complex architectures ensure RPLAs are not only interactive but also adaptive and context-aware, essential for complex role-playing scenarios.



    \item \textbf{Evaluation}:  Evaluating the performance of role-playing models is crucial for assessing their effectiveness and guiding improvements. Commonly, role-playing evaluation involves a complex array of perspectives that extend beyond those applied to traditional dialogue systems, focusing on role consistency, engagement, human-likeness, and proactivity, among others. This complexity necessitates a diverse set of evaluation metrics, encompassing both subjective \textit{human-based} assessments and objective \textit{reference-based}. The advent of LLMs has also spurred the development of \textit{LLM-based evaluation} methods, offering an efficient alternative to conventional human annotation by approximating human judgments. 
    However, given the multifaceted nature of role-playing, no single metric suffices to fully assess their performance. Thus, a composite approach, utilizing multiple metrics in tandem, is essential for a comprehensive evaluation.
    

\end{itemize}

In the following sections, we present a comprehensive survey along with our taxonomy.

%% file: sections/4Benchmarks.tex
\section{Data}


\label{dataset}



\input{figures/data_source}

\subsection{Data Objectives}
Unlike traditional text generation tasks, the capabilities of a role-playing language model are primarily determined by the target dataset it is fitted on. Therefore, the dataset is the most crucial prerequisite for training, categorizing, and testing different role-playing dialogue agents. Commonly, role-playing datasets contain two important components: \textbf{interactions} and \textbf{role-related information}. It's worth noting that we use the term "interactions" instead of the commonly used "conversations" or "dialogues". This is because we believe that the essence of role-playing lies in mimicking the behavior of the role in any scenarios, not just in dialogues. The reason why most current research is limited to the conversation-level is that, compared to other scenarios, the user's actions within a conversation are the easiest to obtain.

In this study, based on the different objectives of the targeted datasets, we categorize role-playing applications into two types: \textbf{Persona-based Role-Playing} (\textbf{P-RP}) and \textbf{Character-based Role-Playing} (\textbf{C-RP}). Generally, P-RP means that the dataset contains coarse-grained role-related information, whereas building C-RP requires fine-grained role-related information. \textit{The classification of granularity of coarse-grained or fine-grained primarily depends on whether the role-related information includes character-level specific background details}.  Datasets lacking such details are deemed coarse-grained, whereas those containing them are considered fine-grained.
Here, we further emphasize the  differences:




\begin{itemize}[leftmargin=*]
    \item \textbf{Persona-based Role-Playing} (P-RP): These role-playing scenarios mimic broad categories of personas, focusing primarily on superficial and common attributes like location and gender. They are expected to display characteristics of specific
groups of people based on given \textit{coarse-gained} personas \cite{Zheng_Zhang_Huang_Mao_2020,dinan2018wizard,DBLP:conf/ijcai/KotturWC17,zhang-etal-2018-personalizing}.
    P-RP is simpler, designed to ensure consistency within a narrower set of persona traits. They are generally less complex and tailored for generic roles that require basic interaction capabilities.

\item \textbf{Character-based Role-Playing} (C-RP): In contrast, C-RP scenarios are crafted to emulate specific characters from various narratives, such as novels, movies or even celebrities. These involves incorporating \textit{fine-grained} character-level personal background information, including
attributes, complex relationships, scene and nuanced psychological states \cite{chen2023large, chen2024roleinteract, tu2024charactereval}. C-RP with language models is inherently complex, aimed at achieving deep, role-specific interactions and are equipped with features such as human-likeness, empathy, and proactivity \cite{zhang2024unveiling}. They are designed to offer a more immersive and engaging user experience.

\end{itemize}

\textbf{Note}, in the evolving landscape of role-playing research, Persona-based Role-Playing (P-RP) is seen as a specific subset within the broader spectrum of Character-based Role-Playing (C-RP). Currently, the focus is increasingly on C-RP, reflecting a shift toward more complex and nuanced scenarios capable of leveraging the sophisticated capabilities of curret LLMs. This trend underscores a growing interest in developing role-playing interactions that offer rich, character-driven experiences. 
We classify different datasets 
according to interaction collection in Figure \ref{data} and present overview of existing datasets
in Table \ref{table:compare-benchmarks}.


\subsection{Persona-based Role-Playing Datasets}


In general, the datasets associated with persona-based data tend to provide personas that are coarse-grained. The collection and construction of appropriate training data have become essential prerequisites for enabling chit-chat dialogue agents to generate persona-specific responses.


\paragraph{Interaction Collection} According to how to collect role-related conversations, we can generally classify the data collection process into two main streams:


\begin{itemize}[leftmargin=*]
    \item \textbf{Employing Crowdsourced Workers}: This method involves hiring crowdworkers to create the corresponding dialogue datasets. Initially, certain personas and related topics are manually defined. Then, crowdworkers engage in dialogues based on these provided personas, resulting in personalized dialogues. The advantage of this approach lies in the guaranteed high quality of the data; however, due to the cost of manual labor, the scale is often limited. Notable examples include the Persona-Chat dataset \cite{zhang-etal-2018-personalizing}, which contains about 10,000 persona dialogues, and the Focus \cite{DBLP:journals/corr/abs-2112-08619} dataset with approximately 14,000 conversations.
    \item \textbf{Extracting from Social Media}: This method typically involves collecting a large volume of user dialogue data from social media platforms and applying specific filtering rules to obtain the final dataset. The strengths of these datasets are that they reflect real-world personalized dialogues and are large in scale. For instance, PersonalDialog \cite{DBLP:journals/corr/abs-1901-09672} and Pchatbot \cite{DBLP:conf/sigir/QianLZGMZLDW21} have respectively gathered over 20 million and 130 million dialogue sessions from Weibo. However, a significant drawback is the difficulty in controlling data quality.
\end{itemize}

In general, there is a trade-off between the above two collection processes: The former provides high-quality, controlled data at a smaller scale, while the latter offers extensive real-world dialogue data, albeit with potential quality control issues.

\paragraph{Role-related Information.}

\input{tables/benchmark}

Role-related information in P-RP datasets is crucial for generating realistic and context-appropriate responses, and it can be categorized into two distinct forms: \textbf{explicit} and \textbf{implicit}. 
Explicit information refers to instances where each interaction is accompanied by a detailed persona, presented either in a natural language format \cite{zhang-etal-2018-personalizing} or a structured format \cite{song2020profile,DBLP:journals/corr/abs-1901-09672}. A prime example of the former, a persona in natural language format, is as follows (sourced from paper \citet{zhang-etal-2018-personalizing}):

\textit{RPGs are my favorite genre.}

 \textit{I also went to school to work with technology.}
 
\textit{The woman who gave birth to me is a physician.}

\textit{I am not a social person.}

\textit{I enjoy working with my hands.}

As for the structured format, an example from \citep{DBLP:journals/corr/abs-1901-09672} is as follows:

\textit{\{Age: Post-90s,}

\textit{Gender: Female,}

\textit{Location: Beijing, }

\textit{Constellation: Aquarius\}}

To the best of our knowledge, most real-world role-playing applications incorporate both formats of role-related information. The key-value structured format is utilized to provide common, fundamental details that are necessary for each role, such as height, weight, and gender. On the other hand, the natural language format is employed to convey unique background information, experiences, and catchphrases specific to the role, which are challenging to encapsulate within the confines of a structured format.

In the context of implicit personas, as referenced in sources such as ~\cite{,DBLP:conf/acl/LiGBSGD16,DBLP:journals/www/ZhangZWZL19}, some might contend that they do not fall within the scope of role-playing, given that they do not provide any role information. Such datasets distinguish dialogues belonging to different roles but do not supply persona information for each role. We still consider these works to be part of the role-playing domain, as the role-related information can essentially be inferred by summarizing the interaction history. We categorize these works under "Implicit persona". Undoubtedly, utilizing these types of datasets for role-playing is a more challenging task, as we must first devise methods to extract accurate role-related information from the interaction history.

\subsection{Character-based Role-Playing Datasets}





As LLMs' comprehension capabilities have improved, both researchers and users have found that coarse-grained personas are no longer sufficient to meet their needs for entertainment and psychological engagement. They have started to use LLMs to `recreate' their favorite characters, a practice known as Character-based Role Playing. Typically, the role-related information provided by Character-based Role-playing can be a comprehensive role description (comprising thousands of tokens), or it could be corpus-level background materials such as novels or narratives. A pioneering effort in this field is the HPD dataset~\cite{chen2023large}, a dataset based on the Harry Potter novels, which is used to train LLMs to align with the Harry Potter.

\paragraph{Interaction Collection} 
Character-based role-playing scenarios involve simulating a broad spectrum of roles, categorized mainly into two categories: \textbf{real world-based} and \textbf{virtual scenario-based}. Real world-based roles often mimic actual celebrities or typical individuals from daily life, while virtual scenario-based roles draw from fictional sources like novels, TV series, video games, and even theoretical constructs such as MBTI personalities \cite{wang2024incharacter}. The diversity of these roles necessitates equally diverse methodologies for data collection, each with its unique set of challenges and solutions:

\begin{itemize}[leftmargin=*]

\item \textbf{LLM as Data Generator}: With the advanced generative capabilities of models like GPT-4, LLMs serve as a primary tool for synthesizing character profiles and dialogues. Related implementations include: 1) Generating complete character profiles from scratch and using these profiles to prompt varied character-based dialogues \cite{wang2023rolellm}; 2) Employing established profiles from resources like MBTI or Wikipedia as a basis for generating personalized dialogues \cite{zhou2023characterglm, chen2024roleinteract,lu2024large}. While effective, LLM-based data generation can introduce biases and unpredictable variations in data quality, necessitating careful manual review and validation \cite{anwar2024foundational}.

    \item \textbf{Extracting from Literary Resources}: This method involves the extraction of role-related conversations and character backgrounds directly from literary sources, particularly for fictional characters depicted in novels and television \cite{chen2023large, tu2024charactereval,zhao2023narrativeplay,wu-etal-2024-role}. However, this approach faces several challenges:
1) Dialogues are often tied to specific scenes, making it complex to delineate the scene of the dialogue and to extract the relevant textual context.
2) Automatically extracting detailed background information about characters is difficult.
3) Identifying multiple statements made by a character within a dialogue round can be complex.
4) Some dialogues involve non-verbal cues and contexts that are difficult to convey textually.
A common solution involves using LLMs, human annotators, or a combination of both. For instance, \citet{chen2023large} utilized a group of professional annotators and the LLM tools to annotate dialogue, attributes, and relationships from the Harry Potter novels. 

Undoubtedly, the language quality of the interactions extracted from this source is exceptional, and they exhibit the highest degree of alignment, given that the characteristics of the roles are originally defined by these works. However, the primary challenge lies in the significant disparity between the dialogue style in the literaries and the daily user-AI dialogue style. This discrepancy often results in the trained models underperforming in interactions with real users.

\item \textbf{Human Role-Playing}: \citet{zhou2023characterglm} and \citet{zhang2024unveiling} hire different crowdworkers who are given specific character profiles to role-play, and then engage in interactions.  This data generation method generally produces high-quality data but lacks diversity and is  costly.

\item \textbf{Unpublished Resources:} All the aforementioned acquisition methods are sourced from formal, published academic papers. However, according to our practical experience, solely relying on these datasets is inadequate to train a product-level role-playing model that fulfills user expectations. Consequently, we present a list of higher-quality role-playing data sources. However, it's important to note: we do not guarantee the legality of these data resources, and the developers should confirm any potential legal risks by themselves. 

The first type of resource is role-play forums, which contain a vast amount of human-human role-playing data. Some well-known forums include Blue Moon\footnote{https://bluemoonroleplaying.com/community/}, NationStates\footnote{https://forum.nationstates.net/}, Aryion\footnote{https://aryion.com/}, Questionable Questing\footnote{https://forum.questionablequesting.com/}, Role-Player\footnote{https://role-player.net/forum/blog.php}, and Spacebattles\footnote{https://forums.spacebattles.com/}. It's important to note that these forums often contain adult-only content, rigorous data cleaning is required before use. The second type is the log of some online role-playing products~\cite{chen2024roleinteract, chen2024compress} such as CharacterAI\footnote{https://character.ai/}. But the use of such data requires dual authorization from both the users and the product developers. The last type is fanfiction communities, such as AO3\footnote{https://archiveofourown.org/}. For some well-known characters like Harry Potter, the volume of fanfiction is thousands of times that of the original work. However, the risk lies in the fact that there are many Out-of-Character situations in fanfiction, as authors will add many personas according to their own preferences.

\end{itemize}

\begin{figure*}[!t]
\vspace{-10pt}
\centering
\includegraphics[width=0.95\linewidth]{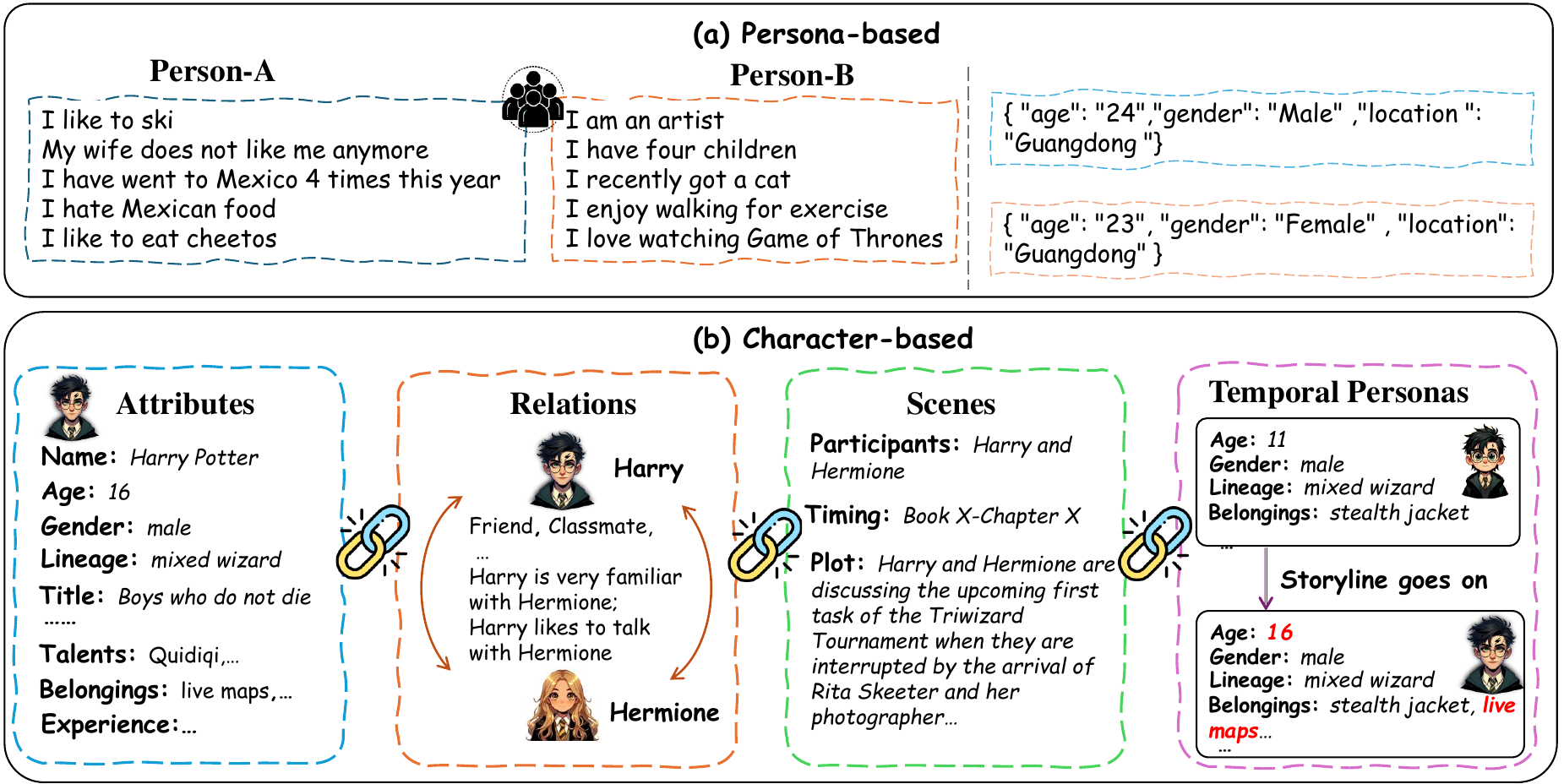}
\caption{Typical (a) persona-based and (b) character-based role-related information in  role-playing datasets. The given examples (a) are from Persona-Chat \cite{zhang-etal-2018-personalizing} and PersonalDialog \cite{DBLP:journals/corr/abs-1901-09672}.
Example (b) 
is collected from HPD \cite{chen2023large}. 
}
\label{fig:finegrained}
\end{figure*}
\paragraph{Role-related Information.} 
Character-based role-playing needs character-level fine-grained personal background information, which encompasses a range of elements, as shown in Figure \ref{fig:finegrained}. Below is a detailed categorization of the typical personal content found within these datasets:


\begin{itemize}[leftmargin=*]
    \item \textbf{Specific Scenes}: This category captures detailed contextual information about the dialogues, including the precise timing, location, and underlying reasons \cite{tu2024charactereval, chen2023large}. Such data is essential for situating the dialogue within a clear narrative frame, thereby providing a more immersive experience.
\item \textbf{Comprehensive Attributes}:  Attributes cover a wide spectrum of personal details about the characters \cite{zhou2023characterglm,tu2024charactereval,wang2023rolellm}, including the \textit{name, gender, personality, age, identity, title, affiliation, quotes, belongings}, etc.
\item \textbf{Complex relationships}: 
Relationship is another key perspective to make role-playing models delve deeper into the social and emotional landscapes of the specific characters \cite{chen2024roleinteract,chen2023large}, which contains
\textit{opponents, familiarity}, and \textit{family details}, etc.
\item \textbf{Temporal Personas}: This aspect acknowledges that characters are not static; their personal information can evolve over time, reflecting developments in the character's storyline or changes in their environment. This temporal dimension allows for the simulation of growth and transformation, providing a richer narrative \cite{shen2024roleeval,chen2023large,ahn2024timechara}.
\end{itemize}

Each category of data plays a critical role in constructing nuanced and responsive character-level role-playing models which can interact in a manner that mirrors human-like complexity and depth.  Note, not all datasets  contain each of the aforementioned elements. For instance, CharacterEval \cite{tu2024charactereval} provides scene and attributes while RoleInteract \cite{chen2024roleinteract} contains scene, attributes and relations.


%% file: figures/data_source.tex
\tikzstyle{my-box}=[
    rectangle,
    draw=hidden-black,
    rounded corners,
    text opacity=1,
    minimum height=1.5em,
    minimum width=5em,
    inner sep=2pt,
    align=center,
    fill opacity=.5,
]
\tikzstyle{cause_leaf}=[my-box, minimum height=1.5em,
    fill=harvestgold!20, text=black, align=left,font=\scriptsize,
    inner xsep=2pt,
    inner ysep=4pt,
]
\tikzstyle{detect_leaf}=[my-box, minimum height=1.5em,
    fill=purple!27, text=black, align=left,font=\scriptsize,
    inner xsep=2pt,
    inner ysep=4pt,
]
\tikzstyle{mitigate_leaf}=[my-box, minimum height=1.5em,
    fill=lightgreen!20, text=black, align=left,font=\scriptsize,
    inner xsep=2pt,
    inner ysep=4pt,
]
\tikzstyle{eval_leaf}=[my-box, minimum height=1.5em,
    fill=pink!10, text=black, align=left,font=\scriptsize,
    inner xsep=2pt,
    inner ysep=4pt,
]
\tikzstyle{sys_leaf}=[my-box, minimum height=1.5em,
    fill=orange!10, text=black, align=left,font=\scriptsize,
    inner xsep=2pt,
    inner ysep=4pt,
]
\begin{figure*}[tp]
    \centering
    \begin{adjustbox}{width=1.0\textwidth}
        \begin{forest}
            forked edges,
            for tree={
                grow=east,
                reversed=true,
                anchor=base west,
                parent anchor=east,
                child anchor=west,
                base=left,
                font=\small,
                rectangle,
                draw=hidden-black,
                rounded corners,
                align=left,
                minimum width=4em,
                edge+={darkgray, line width=1pt},
                s sep=3pt,
                inner xsep=2pt,
                inner ysep=3pt,
                line width=0.8pt,
                ver/.style={rotate=90, child anchor=north, parent anchor=south, anchor=center},
            },
            where level=1{text width=4.1em,font=\scriptsize,}{},
            where level=2{text width=7.2em,font=\scriptsize,}{},
            where level=3{text width=7.2em,font=\scriptsize,}{},
            where level=4{text width=6.4em,font=\scriptsize,}{},
            [
                Data, ver, 
                text=black
                [
                Persona-based ,  text=black
                        [
                            Employing Crowdsourcing \\ Workers,  text=black
                            [
                                {\eg~Persona-Chat \cite{zhang-etal-2018-personalizing}, Focus \cite{DBLP:journals/corr/abs-2112-08619}, KvPI \cite{song2020profile},\\ WOW \cite{dinan2018wizard}, ConvAI2 \cite{dinan2019second}}, detect_leaf, text width=27.2em
                            ]
                        ]
                        [
                            Extracting from    Social \\Media, text=black
                            [
                                {\eg~Pchatbot \cite{DBLP:conf/sigir/QianLZGMZLDW21}, PeDialog \cite{DBLP:journals/corr/abs-1901-09672}, LiveChat \cite{gao2023livechat},\\ P-Ubuntu \cite{li2021dialogue}, Friends-QA \cite{DBLP:conf/sigdial/YangC19}}
                                , detect_leaf, text width=27.2em
                            ]
                        ]
                    ]
                    [
                        Character-based, text=black
                        [
                            LLMs as  Data Generator,  text=black
                            [
                                {\eg~InCharacter \cite{wang2024incharacter}, RoleInteract \cite{chen2024roleinteract}, Rolebench \cite{wang2023rolellm}, \\CharacterDia \cite{zhou2023characterglm}, MBTI-S2Conv \cite{tu2023characterchat}
                                }
                                , detect_leaf, text width=27.2em
                            ]
                        ]
                        [
                            Extracting from \\ Literary Resources,  text=black
                            [
                                {\eg~HPD \cite{chen2023large}, CharacterEval \cite{tu2024charactereval}, RoleEval \cite{wang2023rolellm}, \\CharacterDia \cite{zhou2023characterglm}, Timechara \cite{ahn2024timechara} }
                                , detect_leaf, text width=27.2em
                            ]
                        ]
                        [
                            Unpublished Resources, text=black
                            [
                                {\eg~Dolphin \cite{chen2024compress}, RoleEval \cite{shen2024roleeval}, PIPPA \cite{gosling2023pippa}}
                                , detect_leaf, text width=27.2em
                            ]
                        ]
                        [
                            Human Role-Playing, text=black[{\eg~CharacterDia \cite{zhou2023characterglm}, \citet{zhang2024unveiling}}
                                , detect_leaf, text width=27.2em]  
                            ]
                        ]              
                        ]
        \end{forest}
        \end{adjustbox}
    \caption{The main content flow and categorization of Data Section. }
    \label{data}
    
\end{figure*}

%% file: tables/benchmark.tex
\begin{table*}[t]
\tiny
\centering
\begin{adjustbox}{width=0.98\textwidth}
\begin{tabular}{llcccccccc}
\toprule
&\textbf{ Dataset}  & \textbf{Scene}   & \textbf{Attributes}& \textbf{Relations}& \textbf{Temporal} &\textbf{Source} & \textbf{Usage} &\textbf{Annotation} & \textbf{Language}

\\ \midrule
\multirow{8}{*}{\rotatebox{90}{\textbf{Persona-based}}}& Pchatbot \cite{DBLP:conf/sigir/QianLZGMZLDW21}&$\times$&$\surd$ & $\times$ &$\times$& Weibo & Train \& Test & Rule & Zh \\

& PersonalDialog \cite{DBLP:journals/corr/abs-1901-09672}&$\times$& $\surd$  & $\times$& $\times$& Weibo & Train \& Test & Rule & Zh\\

& KvPI \cite{song2020profile}&$\times$& $\surd$  & $\times$&$\times$& Weibo & Train \& Test & Human & Zh\\

&Personal-Chat \cite{zhang-etal-2018-personalizing} & $\times$& $\surd$  & $\times$& $\times$& Crowdsourcing & Train \& Test & Human & En \\
&RealPersonaChat \cite{yamashita2023realpersonachat} & $\times$& $\surd$  & $\times$& $\times$& Crowdsourcing & Train \& Test & Human & JP \\
    &COMSET \cite{agrawal2023multimodal} &$\surd$ &$\surd$ & $\times$& $\times$& Comics & Train \& Test & Tool-Human & En \\
 &WOW \cite{dinan2018wizard}&$\surd$ &$\surd$  & $\times$& $\times$& Wikipedia &  Train \& Test & Human & En \\
 
& Fri-QA \cite{DBLP:conf/sigdial/YangC19} &$\times$ &$\surd$  & $\surd$& $\times$& TV-Show & Train \& Test & Human & En \\

 & Focus \cite{DBLP:journals/corr/abs-2112-08619}&$\times$ &$\surd$  & $\times$& $\times$& AMT &   Train \& Test & Human & En\\

&UltraChat \cite{ding2023enhancing}&$\surd$ &$\surd$  & $\times$& $\times$& Real-World &Train \& Test & LLM & En \\
&LiveChat \cite{gao2023livechat}&$\times$ &$\surd$  & $\times$& $\times$& Live Videos &Train \& Test & LM-Human & Zh \\
&\citet{cho2023crowd}  & $\times$ & $\surd$& $\times$& $\times$& Crowdsourcing& Train \& Test & Human & KR\\
 \bottomrule
\multirow{13}{*}{\rotatebox{90}{\textbf{Character-based}}}& \textbf{LaMP} \cite{salemi2024lamp}&$\surd$ &$\surd$  & $\times$& $\times$& Existing Corpus & Test & - & En \\

&\textbf{CharacterEval} \cite{tu2024charactereval} & $\surd$ & $\surd$& $\times$& $\times$& Novels & Test & LLM-human & Zh \\

&\textbf{RoleInteract} \cite{chen2024roleinteract} & $\surd$ & $\surd$& $\surd$& $\times$& Multi-Sources & Test & LLM-Human & En-Zh\\
&\textbf{RoleEval} \cite{shen2024roleeval} & $\surd$ & $\surd$& $\surd$& $\surd$& Multi-Sources & Test &LLM-Human & En-Zh\\
&\textbf{TimeChara} \cite{ahn2024timechara} & $\surd$ & $\surd$ & $\surd$ &$\surd$ &Novels & Test &LLM &En  \\
&\textbf{Cross-MR} \cite{yuan2024evaluatingcharacterunderstandinglarge} & $\surd$ & $\surd$ & $\surd$ &$\times$ &Multi-Sources & Test &LLM &En \\
&\textbf{LifeChoice} \cite{xu2024character}& $\surd$ & $\surd$ & $\times$ &$\times$ &Novels & Test &LLM-Human &En \\
&\textbf{PIPPA} \cite{gosling2023pippa} & $\surd$ & $\surd$ & $\times$ &$\times$ &Character.AI & Train &- &En \\
&\textbf{Character100} \cite{wang-etal-2024-characteristic-ai} & $\surd$ & $\surd$& $\times$& $\times$& Novels & Train \& Test & LLM & En \\
&\textbf{HPD} \cite{chen2023large} & $\surd$ & $\surd$& $\surd$& $\surd$& Novels & Train \& Test& LLM-Human & En-Zh\\
&\textbf{RoleBench} \cite{wang2023rolellm} & $\times$ & $\surd$& $\times$& $\times$& Scripts& Train \& Test & LLM & En-Zh\\

&\textbf{CharacterDia} \cite{zhou2023characterglm}& $\surd$ & $\surd$& $\surd$& $\times$& Multi-Sources  & Train \& Test & LLM-Human & Zh\\

&\textbf{ChatHaruhi} \cite{li2023chatharuhi} & $\times$ & $\surd$& $\times$& $\times$& Literary Sources & Train \& Test & LLM & En-Zh\\

&\textbf{MBTI-S2Conv} \cite{tu2023characterchat} & $\times$ & $\surd$& $\times$& $\times$& MBTI & Train \& Test & LLM & En\\

&\textbf{Prodigy} \cite{occhipinti2023prodigy} & $\times$ & $\surd$& $\times$& $\times$&  Literary Sources & Train \& Test & LLM-Human & En\\

&\textbf{Ditto} \cite{lu2024large} & $\times$ & $\surd$& $\times$& $\times$& Open Sources & Train \& Test & LLM & En-Zh\\

&\textbf{Character-LLM} \cite{shao2023character} & $\surd$ & $\surd$& $\times$& $\times$& Wikipedia & Train \& Test & LLM&En \\

\bottomrule
\end{tabular}
\end{adjustbox}
\caption{Overview of existing  datasets for role-playing. }
\vspace{-10pt}
\label{table:compare-benchmarks}

\end{table*}

%% file: sections/6Alignment.tex
\section{Models and Alignment}
\label{alignment}

\input{figures/modeling_alignment}




\input{sections/3Methods}

\subsection{Alignment}
Role-playing hinges on the precise alignment of language models with distinctive character-related information. In other words, alignment plays a crucial role in defining the upper limits of a model's role-playing ability.
Current methodologies for aligning language models with different roles fall into two broad categories: parameter-tuning alignment and parameter-frozen alignment.

\subsubsection{Parameter-Tuning Alignment}

Parameter-Tuning involves adjusting the model's parameters to learn character-specific knowledge. In this manner, common approaches include:
\begin{itemize}[leftmargin=*]
    \item \textbf{Continue-Pretrain}: Help models obtain character-related knowledge, addressing the domain gap between general pre-training and down-stream role-playing. This is essential because generic LLMs lack the nuanced understanding required to portray characters faithfully \cite{chen2023large}. ChatPlug \cite{tian2023chatplug} and MCP
 \cite{DBLP:conf/emnlp/HuangD0M22} train models on targeted literary corpora, capturing subtle narrative cues and character-specific lexicons essential for authentic character representation. To be more specific, ChatPlug  continuly trains Qianwen series \cite{bai2023qwen} on large-scale corpora including common documents and conversation corpus, pursuiting extensive open-world knowledge and foundation abilities, which are used for playing the role of celebrities in the following role-playing scenarios.
 

    \item \textbf{Supervised Fine-tuning (SFT)}: This is the most direct and conventional training approach, involving the concatenation of personal information and conversations for supervised learning. Various techniques \cite{song-etal-2021-bob,zhang-etal-2018-personalizing,10.1145/3477495.3531957} have been employed to enhance the learning of persona information during this phase.
    The core of these supervised methods lies in how to effectively model both role-related information and conversations simultaneously. Notable works include the use of attention routing mechanisms \cite{Zheng_Zhang_Huang_Mao_2020} or memory networks \cite{zhang-etal-2018-personalizing} to integrate both, and 
    employing multiple structures to enhance the model's understanding of both elements \cite{bae-etal-2022-keep, song-etal-2020-generate}. Currently, instruction tuning \cite{wei2022finetuned} has become the mainstream method for fine-tuning  LLMs. During training, specific instructions and character-related data are provided, and the LLM learns through the next token prediction objectives.  Typical works include RoleLLM \cite{wang2023rolellm}, CharacterLLM \cite{shao2023character} and CharacterGLM \cite{zhou2023characterglm}.


    \item \textbf{{Self-Alignment}}: 
    Self-alignment, which is regarded as a new approach that improves weaker LLMs by fine-tuning
it on outputs from a stronger LLM. Inspired by this, CharacterGLM \cite{zhou2023characterglm} and Ditto \cite{lu2024large} employ self-generated data to further encourage LLMs to simulate role-play dialogues. Ditto proposes three steps for role-playing self-alignment: role knowledge collection, dialogue simulation and instruction tuning. It first collects role profiles from open-access knowledge bases such as Wikipedia and then simulates role-playing conversation corpus by conducting a reading comprehension task. At last, Ditto trains the models based on the self-generated datasets to enhance their role-playing abilities.


    \item  \textbf{{Parameter-Efficiency Fine-Tuning (PEFT)}}: Given the vast parameters of current LLMs, training efficiency becomes a critical concern. Techniques like LoRA-tuning \cite{hu2021lora,dettmers2024qlora} selectively train a subset of model parameters, which conserves computational resources. 
    \citet{DBLP:conf/sustainlp/HanGJYZLG23} propose PersonaPKT, which represents each persona as condensing vectors to learn implicit persona-specific features. Such method only require less than  0.1\% trainable parameters of the backbone while maintaining good response generation quality.
    Moreover, \citet{yu2024neeko} employ different LoRA modules to help LLMs imitate multiple characters simultaneously, balancing effectiveness with efficiency. 
    \item \textbf{{Reinforcement Learning}}: RLHF \cite{DBLP:conf/nips/Ouyang0JAWMZASR22, lambert2022illustrating} is a pivotal enhancement approach utilized predominantly after the SFT stage. In the context of role-playing, RLHF-related methods can also help LLMs to refine and align the generated responses more closely with the intended character traits and behaviors. 
\citet{shea-yu-2023-building} utilizes offline RL strategies to improve the persona consistency.
    Similarly, COMEDY \cite{chen2024compress}
    utilizes GPT-4 to construct memory-based personal responses and memory-against personal responses, forming positive-negative pairs, and then employ DPO \cite{rafailov2023direct} strategies for aligning LLMs to generate more coherent memory-based personalized responses. 

    However, based on our practical engineering experience, we've found that 1) Reinforcement learning approaches, when using a reward model constructed through in-context learning, generally cannot exceed their inherent role-playing capabilities. 2) The task of annotating high-quality preference data for role-playing is significantly more challenging than for a generic assistant, as it necessitates a deep understanding of the specific character to accurately annotate preferences. For example, during the annotation of the HPD datasets~\cite{chen2023large}, the authors enlisted the help of five avid Harry Potter fans to annotate Harry Potter's behavior.
    
\end{itemize}


\subsubsection{Parameter-Frozen Alignment}

The parameter-frozen alignment approaches in role-playing offer a versatile framework to adapt to new roles without extensive retraining of the model's parameters. These methods focus on utilizing existing model capabilities and enhancing them through strategic use of external data and contextual prompts. 

\begin{itemize}[leftmargin=*]
    \item \textbf{In-Context Learning (ICL) Prompting}: ICL-based approaches leverage the LLM's inherent ability to contextualize and adapt its responses based on provided prompts and examples within the inputs \cite{wei2022chain, brown2020language}.  To simulate the behavior of specific roles, ICL prompting is the simplest approach. Typically,  filling with role attributes, relations, task requirements within ICL, current LLMs can adapt to different roles swiftly \cite{park2023generative, shao2023character, tu2024charactereval}. This method is highly effective for rapid deployment across varied characters \cite{zhao2023narrativeplay}, making it ideal for scenarios where models need to switch roles or adapt to new narratives. 
    \citet{park2023generative} even assign identities to multiple agents via ICL, make them simulating different roles in ``WestWorld'' sandbox game.

    


    \item \textbf{Retrieval Augmented Generation (RAG)}: RAG \cite{shuster2021retrieval, li2022survey} enhances role-playing by dynamically retrieving data from external databases before response generation. This method addresses the internal knowledge gaps of models about specific characters \cite{liu-etal-2023-recap, zhao2023narrativeplay, salemi2024lamp}, reducing hallucinations—factually incorrect but plausible responses. By grounding responses in verifiable data, RAG enhances character consistency and enriches dialogues with precise, character-specific details.

\end{itemize}

\subsubsection{Summary and Discussion.} Parameter-tuning alignment offers precise control over role-playing models behaviors, enabling deep customization and high specificity in character dialogue generation. However, they require significant computational resources, high-quality training data
and risk overfitting, potentially reducing the model's generalizability.  

Conversely, parameter-frozen methods like In-context Learning and Retrieval Augmented Generation provide flexibility and scalability, allowing easy adaptation to new characters without extensive retraining. While reducing potential model biases, these methods depend heavily on abilities of LLMs and  the quality and availability of external data, 
which can also introduce inaccuracies and result in hallucinations if not properly managed. Both approaches thus necessitate careful design to ensure the reliability and relevance of dialogue outputs.

%% file: figures/modeling_alignment.tex
\tikzstyle{my-box}=[
    rectangle,
    draw=darkpastelgreen,
    rounded corners,
    text opacity=1,
    minimum height=1.5em,
    minimum width=5em,
    inner sep=2pt,
    align=center,
    fill opacity=.5,
]
\tikzstyle{cause_leaf}=[my-box, minimum height=1.5em,
    fill=darkpastelgreen!20, text=black, align=left,font=\scriptsize,
    inner xsep=2pt,
    inner ysep=4pt,
]
\tikzstyle{detect_leaf}=[my-box, minimum height=1.5em,
    fill=cyan!20, text=black, align=left,font=\scriptsize,
    inner xsep=2pt,
    inner ysep=4pt,
]
\tikzstyle{mitigate_leaf}=[my-box, minimum height=1.5em,
    fill=darkpastelgreen!20, text=black, align=left,font=\scriptsize,
    inner xsep=2pt,
    inner ysep=4pt,
]
\tikzstyle{eval_leaf}=[my-box, minimum height=1.5em,
    fill=pink!10, text=black, align=left,font=\scriptsize,
    inner xsep=2pt,
    inner ysep=4pt,
]
\tikzstyle{sys_leaf}=[my-box, minimum height=1.5em,
    fill=orange!10, text=black, align=left,font=\scriptsize,
    inner xsep=2pt,
    inner ysep=4pt,
]
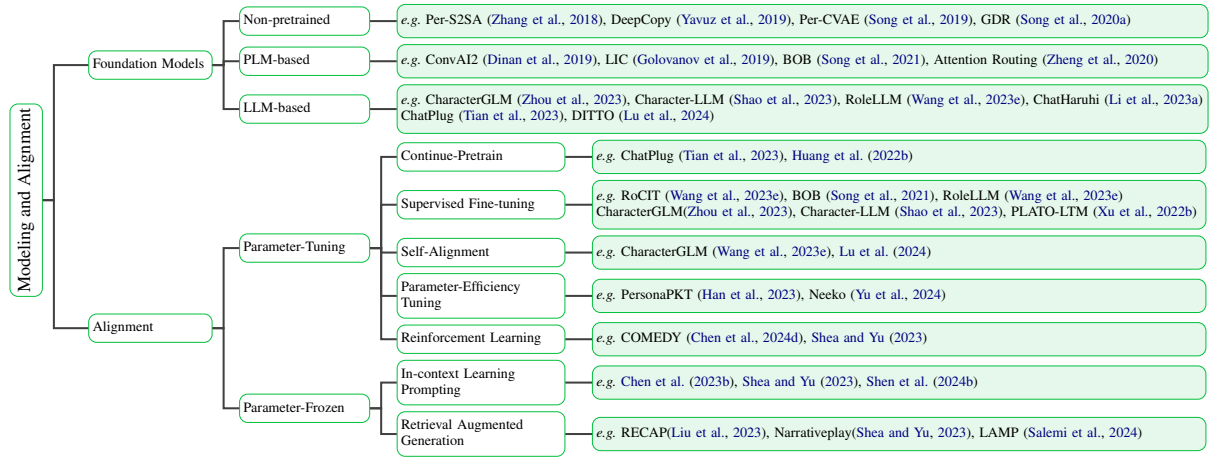
\begin{figure*}[]
    \centering
    \resizebox{\textwidth}{!}{
        \begin{forest}
            forked edges,
            for tree={
                grow=east,
                reversed=true,
                anchor=base west,
                parent anchor=east,
                child anchor=west,
                base=left,
                font=\small,
                rectangle,
                draw=darkpastelgreen,
                rounded corners,
                align=left,
                minimum width=4em,
                edge+={darkgray, line width=1pt},
                s sep=3pt,
                inner xsep=2pt,
                inner ysep=3pt,
                ver/.style={rotate=90, child anchor=north, parent anchor=south, anchor=center},
            },
            where level=1{text width=5.2em,font=\scriptsize,}{},
            where level=2{text width=5.5em,font=\scriptsize,}{},
            where level=3{text width=7.2em,font=\scriptsize,}{},
            where level=4{text width=6.4em,font=\scriptsize,}{},
            [
                Modeling and Alignment, ver, color=darkpastelgreen!100, 
                text=black 
                [
                    Foundation Models, color=darkpastelgreen!100,  text=black[
                    Non-pretrained, color=darkpastelgreen!100,   text=black
                        [
                        {\eg~Per-S2SA \cite{zhang-etal-2018-personalizing}, DeepCopy \cite{yavuz-etal-2019-deepcopy}, Per-CVAE \cite{song2019exploiting}, GDR \cite{song-etal-2020-generate}}
                            , cause_leaf, text width=36.2em
                        ]
                        ]
                    [
                        PLM-based, color=darkpastelgreen!100,  text=black
                        [
                        {\eg~ConvAI2 \cite{dinan2019second}, LIC \cite{golovanov-etal-2019-large}, BOB \cite{song-etal-2021-bob}, Attention Routing \cite{Zheng_Zhang_Huang_Mao_2020}}
                        ,  cause_leaf, text width=36.2em
                        ]
                    ]
                    [
                        LLM-based, color=darkpastelgreen!100, text=black
                        [
                                {\eg~CharacterGLM \cite{zhou2023characterglm}, Character-LLM \cite{shao2023character}, RoleLLM \cite{wang2023rolellm}, ChatHaruhi \cite{li2023chatharuhi}\\ ChatPlug \cite{tian2023chatplug}, 
                                DITTO \cite{lu2024large}}
                                , cause_leaf, text width=36.2em
                        ]
                    ]
                ]
                    [
                    Alignment, color=darkpastelgreen!100,  text=black[
                        Parameter-Tuning, color=darkpastelgreen!100, text=black
                        [
                            Continue-Pretrain, color=darkpastelgreen!100, text=black
                            [
                             {\eg~ChatPlug \cite{tian2023chatplug}, \citet{DBLP:conf/emnlp/HuangD0M22}}, mitigate_leaf, text width=27.3em
                            ]
                        ]
                        [
                            Supervised Fine-tuning , color=darkpastelgreen!100, text=black
                            [
                            {\eg~RoCIT \cite{wang2023rolellm},
                            BOB \cite{song-etal-2021-bob}, RoleLLM \cite{wang2023rolellm}\\CharacterGLM\cite{zhou2023characterglm}, Character-LLM \cite{shao2023character}, PLATO-LTM \cite{DBLP:conf/acl/XuGWNW0W22}}, mitigate_leaf, text width=27.3em
                            ]
                        ]
                        [
                            Self-Alignment, color=darkpastelgreen!100,  text=black
                           [
                           {\eg~CharacterGLM \cite{wang2023rolellm},
                           \citet{lu2024large}}, mitigate_leaf, text width=27.3em
                           ]
                        ]
                        [
                            Parameter-Efficiency \\ Tuning, color=darkpastelgreen!100, text=black
                            [
                            {\eg~PersonaPKT \cite{DBLP:conf/sustainlp/HanGJYZLG23},
                            Neeko \cite{yu2024neeko}}, mitigate_leaf, text width=27.3em
                            ]
                        ]
                        [
                             Reinforcement Learning , color=darkpastelgreen!100,  text=black
                            [
                            {\eg~COMEDY \cite{chen2024compress}, \citet{shea-yu-2023-building}}, mitigate_leaf, text width=27.3em]
                        ]
                    ]
                    [
                        Parameter-Frozen, color=darkpastelgreen!100,  text=black
                        [
                            In-context Learning \\Prompting, color=darkpastelgreen!100,  text=black
                        [
                        {\eg~\citet{chen2023large}, \citet{shea-yu-2023-building}, \citet{shen2024roleeval}}, mitigate_leaf, text width=27.3em
                        ]
                        ]
                        [
                           Retrieval  Augmented\\Generation, color=darkpastelgreen!100,  text=black
                    [{\eg~RECAP\cite{liu-etal-2023-recap}, Narrativeplay\cite{shea-yu-2023-building}, LAMP \cite{salemi2024lamp} }, mitigate_leaf, text width=27.3em]
                    ]
                    ]
                    ]
                    ]       
        \end{forest}
        } 
    \caption{The main content flow and categorization of Models and Alignment. }
    \label{alignment1}
\end{figure*}

%% file: sections/3Methods.tex
\subsection{Foundation Models}

\input{figures/foundation_model}
\label{Models}

Foundation models are critical in setting the base capabilities of role-playing models. As the underlying architectures, they determine the lower bounds of performance and sophistication achievable in role-playing scenarios. 
The development of foundation models and architectures for role-playing can be viewed as a progressive evolution across three distinct stages: \textbf{non-pretrained model}, \textbf{PLM}, and \textbf{LLM}. Each stage represents a significant shift in backbone selection for role-playing models.



\paragraph{Non-pretrained models and PLM.}

The earliest stage in the development of role-playing models involved non-pretrained architectures. These models are crafted from scratch, tailored to specific tasks without the benefit of large-scale, pre-trained data. Early models often utilize bespoke designs such as memory networks or custom transformers, which are specifically engineered to handle the storage and embedding-based fusion of personal information for effective role-playing \cite{zhang-etal-2018-personalizing,JiangPersonaLLMIT,DBLP:conf/ijcai/KotturWC17,DBLP:conf/acl/LiGBSGD16}. These architectures provided highly specialized solutions that were adept within specific contexts but lacked the generalizability and scalability offered by later developments.
The shift to PLMs like BERT mark a substantial enhancement in foundational capabilities \cite{kenton2019bert, raffel2023exploring, Radford2019LanguageMA}.  
These models leverage extensive pre-trained data, enhancing their ability to understand context and generate text, yet they still face limitations in fully grasping role-specific nuances. To overcome these challenges, researchers deploy innovative strategies 
like contrastive learning \cite{DBLP:conf/emnlp/HuangD0M22}, bert-over-bert decouples learning \cite{song-etal-2021-bob} and attention-based fusion mechanisms \cite{Zheng_Zhang_Huang_Mao_2020} improve the integration of personal and dialogue data, enhancing role-playing functionalities.


\paragraph{LLM.}
The current frontier in role-playing development is characterized by LLMs such as GPT-4, which boast an unprecedented scale in parameters and pre-training. Such LLMs offer profound improvements in understanding and generating text, capable of maintaining coherent and contextually rich personal dialogues even with minimal prompting.
The architecture of these models has largely standardized around the decoder-only framework.
Most LLM-based works \cite{chen2023large, tu2024charactereval, ahn2024timechara} customize various characters by configuring their personal background information in prompts, aiming at mimicking the specific role.
Currently, several role-playing specific LLMs are developed to facilitate future research through instruction-tuning, such as CharacterGLM \cite{zhou2023characterglm}, Xingye\footnote{https://www.xingyeai.com/},  Xingchen\footnote{https://tongyi.aliyun.com/}, Index~\cite{Index}, and Baichuan-Character\footnote{https://npc.baichuan-ai.com/index}.

Taking CharacterGLM as an example, let's explore how LLMs can be optimized for role-playing support:  The process begins with collecting character-related training corpus, where detailed character profiles are developed and utilized to engage in dialogues through either human interactions or LLMs, creating a rich dataset that captures the nuances of character-specific conversations. Following data collection, the next phase is instruction tuning, where the character profiles and accumulated dialogue data are organized into structured instructions. This stage also could incorporate the use of diverse prompts for data augmentation, enhancing the model’s ability to generate varied and contextually appropriate responses. 
The final but optional step,  employs self-alignment—using outputs from advanced models for further training—and human feedback to refine and ensure character consistency. This comprehensive approach ensures that LLMs effectively embody and maintain character traits in role-playing scenarios.

Each stage builds upon the previous, with advancements addressing the limitations of earlier models and opening new possibilities for more complex and engaging role-playing interactions.

Beyond the methodologies outlined in public papers, we wish to underscore the influence of pre-training corpora on role-play tasks. From our pre-training experience, a corpus beneficial for training a generic assistant like ChatGPT may not necessarily aid a role-playing task, and the converse is also true. This is understandable, as a generic assistant requires data abundant in `real' world knowledge (such as news, wiki) or data necessitating complex reasoning (like math, code). Novels, particularly those with a worldview divergent from reality, can induce serious knowledge hallucination issues. Role-playing often presupposes a scenario distinct from the real world, where a certain degree of reasonable knowledge hallucination is encouraged. Therefore, to our knowledge, a crucial step in pre-training an effective foundation model for role-playing involves incorporating a substantial amount of novels into the pretraining corpus, especially those with a worldview distinct from reality.


%% file: sections/6System_Architecture.tex
\section{Agent Architecture}
\label{architecture}

\input{figures/system_architecture}

\begin{figure}[!t]
\vspace{-10pt}
\centering
\includegraphics[width=0.95\linewidth]{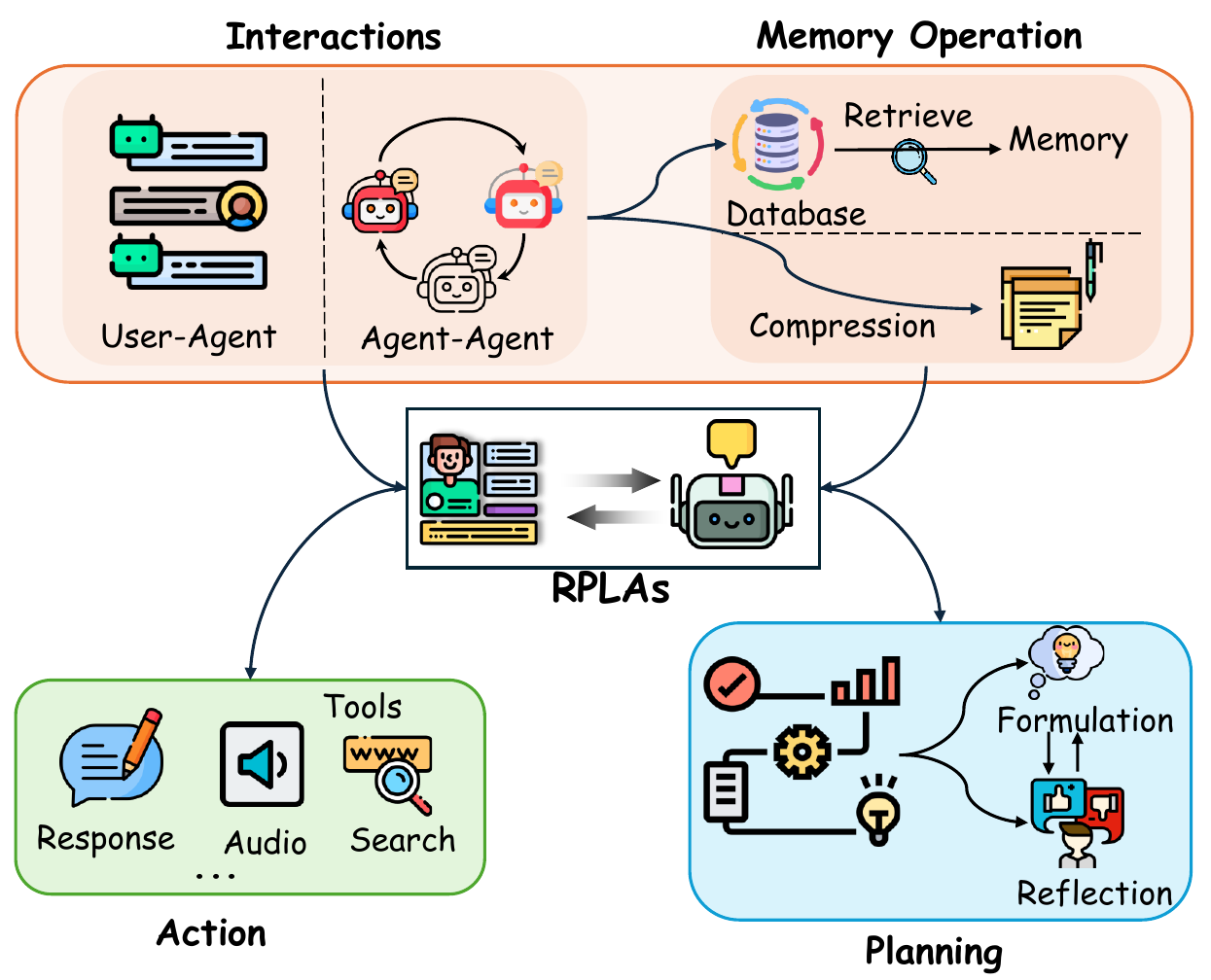}
\caption{Typical system architecture of RPLAs. 
}
\label{fig:rplas}
\vspace{-15pt}
\end{figure}

Building on the foundation of role-playing with large language  models, Role-Playing Language Agents (RPLAs) take this concept further by incorporating interactive and autonomous behaviors \cite{wang2023humanoid, park2023generative, tang2024enhancing, xi2023rise}. These agents not only embody specific characters but also engage in complex scenarios, making decisions and responding in ways that align with their designated roles. Such RPLAs are used to solve different challenging but interesting tasks not limited to original dialogue response generation.
For instance, \citet{chen2024hollmwoodunleashingcreativitylarge} present HOLLMWOOD, which design Writer, Editor
and Actors RPLAs, simulating the human  creative process for screenwriting tasks.
The whole architecture of RPLAs often involves multiple modules that work in tandem. In this section, we introduce three important modules beyond the foundation response functionality in building well-performed RPLAs: \textbf{memory}, \textbf{planning} and \textbf{action}.




\subsection{Memory} RPLAs often operate in environments that require them to remember and synthesize information over time, making memory modules \cite{DBLP:conf/acl/CaoBF0T22} an essential component of their architecture. 
In the RPLA context, memory mainly consists of two source types: 

\begin{itemize}[leftmargin=*]
    \item \textbf{User-Agent Interactions}: This type is important in building user-centric RPLAs, which requires maintaining user related memories for generating consistent personalized responses in the ongoing interaction \cite{bae-etal-2022-keep, DBLP:conf/acl/XuGWNW0W22, chen2024compress}. 
    Specifically, within online role-playing dialogue platforms, the integration of long-term memory becomes crucial in sustaining user engagement and enhancing the immersive experience. In such environments, an RPLA equipped with long-term memory can significantly enrich the role-playing experience by adapting its responses based on the accumulated history of a user’s choices and interactions.
    \citet{xu2024hallucination} propose a long-term memory mechanism to extract and update long-term persona memories from use-bot interactions, enhancing the long-term personalized response consistency. Following this, \citet{bae-etal-2022-keep} further incorporate the memory operator to control the update or ignore of fine-grained memories. 
    



    \item \textbf{Agent-Agent Interactions}: 
    In scenarios involving multiple RPLAs, this type of memory is essential for managing interactions between different agents. It supports scenarios where RPLAs must collaborate or compete within complex environments, such as multiplayer simulation games or interactive narratives. Memory in this context includes recording the outcomes of past interactions, thoughts and actions, which 
    influence future strategies and decisions. This dynamic memory use allows RPLAs to adapt their behaviors based on previous experiences with other agents, fostering a more nuanced and strategic interaction framework that evolves over time.
Generative Agents  \citet{park2023generative} and Humanoid Agents \cite{wang2023humanoid} create a virtual role-playing environment, where an RPLA can remember past alliances or conflicts with other agents, using this information to inform future decisions and interactions.

\end{itemize}





Given different memory forms from previous interactions, there are generally two approaches for integrating them in RPLA systems:

\begin{itemize}[leftmargin=*]
    \item \textbf{Retrieval-based:} This approach is widely-used in current RPLAs, which involve maintaining a database that stores useful information from previous interactions \cite{park2023generative, zhong2024memorybank, DBLP:conf/acl/XuGWNW0W22}. Then facing the ongoing interaction,  a retrieval module like sentence-embedding models \cite{gao2021simcse, chen2023alleviating} fetches the  most relevant information from this database  based on the current context or recent interactions, helping the agent craft appropriate personalized responses  or actions akin to the assigned character. \citet{park2023generative}
retrieve relevant memories to help RPLAs make plans and decisions.    
    Also, several memory management strategies like Memory Operator \cite{bae-etal-2022-keep} are used to ignore useless information and keep the memory base up-to-date.
    While retrieval-based memory offers clear benefits in enhancing RPLA interactions, it also poses challenges: 1)  It heavily depends on the precision of the retrieval model; if these are not capable of accurately identifying relevant information, the coherence and appropriateness of responses can suffer; 2) It require substantial storage to maintain large corpus of past interactions, leading to increased storage costs and demanding efficient data management strategies to ensure quick and effective access to required information. 
    \item \textbf{Compressive-based:} This innovative approach addresses some limitations of retrieval-based memory by internalizing and condensing past information into a compact form, eliminating the need for extensive external databases. Compressive-based memory improves persona consistency by continuously updating and compressing historical data, which allows RPLAs to keep their responses current and relevant, like COMEDY \cite{chen2024compress} and ReSummarize \cite{wang2023recursively}.
    COMEDY employ the ``compress over compress'' idea: it first summarizes each dialogue session into session-level memories, and then condenses them into a final compressive memory. Such method doesn't rely on any sentence-embedding model as retriever and database.
   ReSummarize recursive summarizes the personalized memories from past sessions to keep the up-to-date memories.
    This method enhances the storage efficiency and reduces the dependency on large-scale data retrieval, although it sometimes sacrifice detail for compactness.
\end{itemize}

\subsection{Planning} In the realm of RPLAs, while foundational conversational capabilities are crucial, certain scenarios demand additional capacities such as advanced planning. Let LLMs simulate the human behavior in virtual environments could be a typical example, where strategic planning significantly enhances the role-playing experience \cite{wang2023humanoid, park2023generative}. 
Typically,  planning in RPLAs comprises two stages: plan formulation and plan reflection.

\begin{itemize}[leftmargin=*]
    \item \textbf{Plan Formulation}:   In role-playing contexts, the formulation stage involves setting objectives that are consistent with the character's motivations and the narrative's demands \cite{park2023generative}. Agents analyze the current scenario, predict possible future states, and devise a sequence of actions that will effectively portray their role. This might involve choosing to form alliances, solve puzzles, or navigate through challenges based on the storyline \cite{wang2023unleashing}. \citet{shen2024decisionmaking} prove that GPT4-level LLMs could make different plans according to their assigned MBTIs. Humanoid Agents \cite{wang2023humanoid} further let RPLAs adapt their daily activities with other agents
    through planning.
    
    \item \textbf{Plan Reflection}: After executing a plan, reflection allows RPLAs to assess their actions' effectiveness in the context of the role-play. This might involve introspection on whether the chosen actions adequately advanced the narrative or stayed true to the character's development. In simulations of human-like agents environments, \citet{park2023generative} and \citet{wang2023humanoid} utilize the feedback  mechanisms 
    to help RPLAs refine future planning. Take a simple example, if the agent's actions in a simulated environment lead to unforeseen consequences, reflecting on these outcomes helps the agent adjust its strategy to better fit the narrative and user expectations in future scenarios.
\end{itemize}

By integrating such sophisticated planning and reflection capabilities, RPLAs can offer more dynamic and engaging experiences in role-play environments. These agents not only react to the unfolding story but actively contribute to its progression, making them integral players in shaping how narratives unfold. 

\subsection{Action } 
Agent actions are the culmination of prior planning, memory utilization, and interactions. While the most recognizable form of action is generating responses aligned with role-play, actions within RPLAs extend beyond mere conversations. Tool use \cite{schick2024toolformer} as an another critical component of role-play actions, alongside a brief discussion on the potential for embodied actions in role-playing scenarios.
For example, 
tools like search-related APIs \cite{salemi2024lamp, wang2023large, zhao2023narrativeplay} enable RPLAs to fetch and incorporate relevant character knowledge into their responses, enriching the dialogue with contextually appropriate content. This is particularly vital in scenarios requiring deep domain knowledge or historical accuracy, where tools can provide the necessary data to support the agent's role accurately.


An emerging yet underexplored area in role-playing actions is embodied actions, where agents interact more tangibly within their environments. This involves physical interactions in virtual or augmented realities, representing a promising frontier for developing more immersive role-playing experiences. Although current role-playing applications primarily focus on dialogue and tool use, the potential for incorporating embodied actions offers exciting prospects for future advancements in RPLA interactivity and realism.


\subsection{Summary and Discussion} In RPLAs development, the integration of memory, planning, and action modules is critical for delivering rich, interactive experiences. Current memory usage, both retrieval-based and compressive-based, offers strengths in contextual continuity and processing efficiency, respectively, yet faces challenges like high storage costs and potential loss of detail. Planning modules are essential for strategic decision-making within complex narratives but often struggle with transparency and robustness in dynamic scenarios. Action modules, encompassing dialogue and tool use, are well-developed for enhancing interactions, though the integration of tools needs careful management to avoid detracting from the user experience. Looking forward, advancements in these areas could include more nuanced memory retention, improved planning algorithms for better adaptability and transparency, and more integrated tool use, particularly with the potential expansion into embodied actions.

%% file: figures/system_architecture.tex
\tikzstyle{my-box}=[
    rectangle,
    draw=orange,
    rounded corners,
    text opacity=1,
    minimum height=1.5em,
    minimum width=5em,
    inner sep=2pt,
    align=center,
    fill opacity=.5,
]
\tikzstyle{cause_leaf}=[my-box, minimum height=1.5em,
    fill=harvestgold!20, text=black, align=left,font=\scriptsize,
    inner xsep=2pt,
    inner ysep=4pt,
]
\tikzstyle{detect_leaf}=[my-box, minimum height=1.5em,
    fill=cyan!20, text=black, align=left,font=\scriptsize,
    inner xsep=2pt,
    inner ysep=4pt,
]
\tikzstyle{mitigate_leaf}=[my-box, minimum height=1.5em,
    fill=lightgreen!20, text=black, align=left,font=\scriptsize,
    inner xsep=2pt,
    inner ysep=4pt,
]
\tikzstyle{eval_leaf}=[my-box, minimum height=1.5em,
    fill=pink!10, text=black, align=left,font=\scriptsize,
    inner xsep=2pt,
    inner ysep=4pt,
]
\tikzstyle{sys_leaf}=[my-box, minimum height=1.5em,
    fill=orange!10, text=black, align=left,font=\scriptsize,
    inner xsep=2pt,
    inner ysep=4pt,
]
\begin{figure*}[]
    \centering
    \resizebox{\textwidth}{!}{
        \begin{forest}
            forked edges,
            for tree={
                grow=east,
                reversed=true,
                anchor=base west,
                parent anchor=east,
                child anchor=west,
                base=left,
                font=\small,
                rectangle,
                draw=hidden-draw,
                rounded corners,
                align=left,
                minimum width=4em,
                edge+={darkgray, line width=1pt},
                s sep=3pt,
                inner xsep=2pt,
                inner ysep=3pt,
                ver/.style={rotate=90, child anchor=north, parent anchor=south, anchor=center},
            },
            where level=1{text width=4.0em,font=\scriptsize,}{},
            where level=2{text width=7.5em,font=\scriptsize,}{},
            where level=3{text width=7.2em,font=\scriptsize,}{},
            where level=4{text width=6.4em,font=\scriptsize,}{},
            [
                Agent  Architecture, ver, color=orange!100, 
                text=black
            [
           Memory, color=orange!100,  text=black
            [
            Memory Sources, color=orange!100,  text=black[
            User-Agent Interaction, color=orange!100,  text=black[
              {\eg~\citet{DBLP:conf/acl/XuGWNW0W22}, \citet{JiangPersonaLLMIT}, RoleInteract \cite{chen2024roleinteract}}, sys_leaf, text width=18em
              ]
              ]
               [Agent-Agent Interaction, color=orange!100,  text=black[
              {\eg~\citet{maas2023infinity}, Generative Agent \cite{park2023generative}}, sys_leaf, text width=18em
              ]
              ]         
            ]
            [
            Memory Usage, color=orange!100,  text=black[
            Retrieval-based, color=orange!100,  text=black[
              {\eg~PLATO-LTM \cite{DBLP:conf/acl/XuGWNW0W22}, \citet{bae-etal-2022-keep} }, sys_leaf, text width=18em
              ]
              ]
               [Compressive-based, color=orange!100,  text=black[
              {\eg~COMEDY \cite{chen2024compress}, \citet{wang2023recursively}}, sys_leaf, text width=18em
              ]
              ]         
            ]
            ]
            [
             Planning, color=orange!100, text=black[
              Planning Formulation, color=orange!100, text=black
              [
              {\eg~\citet{dasgupta2023language}, \citet{wang2023recursively}, Inner Monologue \cite{huang2022inner}}, sys_leaf, text width=26.7em
              ]
             ]
             [
              Planning Reflection, color=orange!100,  text=black
              [
              {\eg~Generative Agent \cite{park2023generative}, MORTISE \cite{tang2024enhancing}, \citet{wu2022ai}}, sys_leaf, text width=26.7em
              ]
             ]
             ]
              [
             Action, color=orange!100,  text=black[
             {\eg~RoleLLM \cite{wang2023rolellm}, Humanoid Agents \cite{wang2023humanoid}, \citet{shen2024decisionmaking}}, sys_leaf, text width=35.8em
             ]
             ]
            ]
        \end{forest}
    }
    \caption{The main content flow and categorization of Agent Architecture. }
    \label{architecture}
\end{figure*}

%% file: figures/evaluation.tex
\tikzstyle{my-box}=[
    rectangle,
    draw=hidden-draw,
    rounded corners,
    text opacity=1,
    minimum height=1.5em,
    minimum width=5em,
    inner sep=2pt,
    align=center,
    fill opacity=.5,
]
\tikzstyle{cause_leaf}=[my-box, minimum height=1.5em,
    fill=harvestgold!20, text=black, align=left,font=\scriptsize,
    inner xsep=2pt,
    inner ysep=4pt,
]
\tikzstyle{detect_leaf}=[my-box, minimum height=1.5em,
    fill=cyan!20, text=black, align=left,font=\scriptsize,
    inner xsep=2pt,
    inner ysep=4pt,
]
\tikzstyle{mitigate_leaf}=[my-box, minimum height=1.5em,
    fill=lightgreen!20, text=black, align=left,font=\scriptsize,
    inner xsep=2pt,
    inner ysep=4pt,
]
\tikzstyle{eval_leaf}=[my-box, minimum height=1.5em,
    fill=pink!10, text=black, align=left,font=\scriptsize,
    inner xsep=2pt,
    inner ysep=4pt,
]
\tikzstyle{sys_leaf}=[my-box, minimum height=1.5em,
    fill=orange!10, text=black, align=left,font=\scriptsize,
    inner xsep=2pt,
    inner ysep=4pt,
]
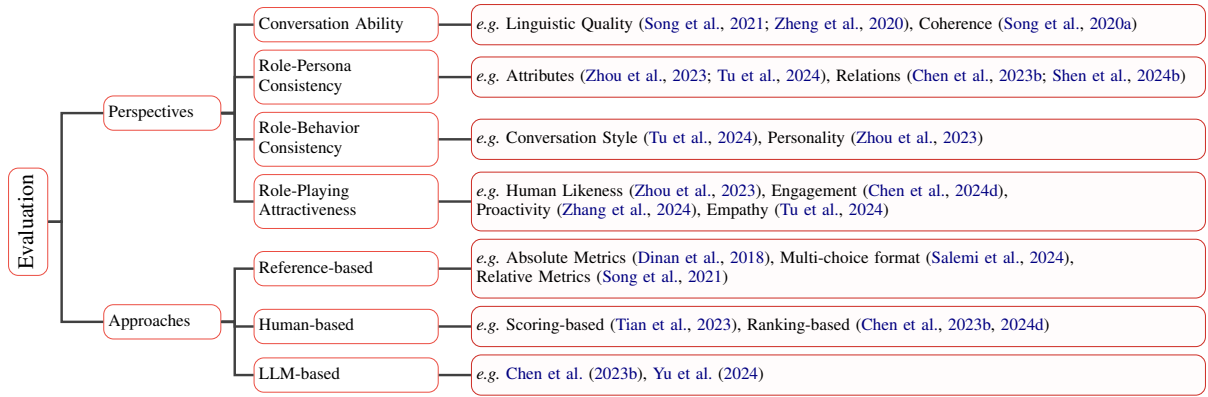
\begin{figure*}[]
    \centering
    \begin{adjustbox}{width=0.99\textwidth}
        \begin{forest}
            forked edges,
            for tree={
                grow=east,
                reversed=true,
                anchor=base west,
                parent anchor=east,
                child anchor=west,
                base=left,
                font=\small,
                rectangle,
                draw=hidden-draw,
                rounded corners,
                align=left,
                minimum width=4em,
                edge+={darkgray, line width=1pt},
                s sep=3pt,
                inner xsep=2pt,
                inner ysep=3pt,
                ver/.style={rotate=90, child anchor=north, parent anchor=south, anchor=center},
            },
            where level=1{text width=4.0em,font=\scriptsize,}{},
            where level=2{text width=6.5em,font=\scriptsize,}{},
            where level=3{text width=7.2em,font=\scriptsize,}{},
            where level=4{text width=6.4em,font=\scriptsize,}{},
            [
                Evaluation, ver, color=carminepink!100, 
                text=black        
            [
            Perspectives, color=carminepink!100, text=black
            [
            Conversation Ability, color=carminepink!100, text=black[
              {\eg~Linguistic Quality \cite{song-etal-2021-bob,Zheng_Zhang_Huang_Mao_2020}, Coherence \cite{song-etal-2020-generate}}, eval_leaf, text width=27em
              ]
            ]
            [
            Role-Persona \\ Consistency, color=carminepink!100,  text=black[
              {\eg~Attributes \cite{zhou2023characterglm,tu2024charactereval}, Relations \cite{chen2023large,shen2024roleeval}}, eval_leaf, text width=27em
              ]
            ]
            [
            Role-Behavior \\ Consistency, color=carminepink!100, text=black[
              {\eg~Conversation Style \cite{tu2024charactereval}, Personality \cite{zhou2023characterglm}}, eval_leaf, text width=27em
              ]
            ]
            [
            Role-Playing \\ Attractiveness, color=carminepink!100,  text=black[
              {\eg~Human Likeness \cite{zhou2023characterglm}, Engagement \cite{chen2024compress}, \\Proactivity \cite{zhang2024unveiling}, Empathy \cite{tu2024charactereval}}, eval_leaf, text width=27em
              ]
            ]
            ]
            [
            Approaches, color=carminepink!100,  text=black
            [
            Reference-based, color=carminepink!100, text=black[
              {\eg~Absolute Metrics \cite{dinan2018wizard}, Multi-choice format \cite{salemi2024lamp}, \\Relative Metrics \cite{song-etal-2021-bob}}, eval_leaf, text width=27em
              ]
            ]
            [
            Human-based, color=carminepink!100, text=black[
              {\eg~Scoring-based \cite{tian2023chatplug}, Ranking-based \cite{chen2023large, chen2024compress}}, eval_leaf, text width=27em
              ]
            ]
            [
            LLM-based, color=carminepink!100, text=black[
              {\eg~\citet{chen2023large}, \citet{yu2024neeko}}, eval_leaf, text width=27em]
            ]
            ]
            ]
        \end{forest}
    \end{adjustbox}
    \caption{The main content flow and categorization of Evaluation Section. }
    \label{categorization_of_survey}
\end{figure*}

%% file: sections/5Evaluation.tex
\section{Evaluation}
\label{evaluation}

Evaluating role-playing models is essential to ensure their effectiveness and realism in simulated environments. Unlike traditional chatbots or nlp tasks, role-playing require a more sophisticated set of evaluation dimensions and metrics due to their unique ability to emulate human-like interactions based on specific characters and contexts. In this section, we will first introduce various perspectives of role-playing evaluation, with a particular emphasis on its differences and connections with generic assistant evaluation. Following this, we will present several mainstream evaluation methods, primarily including reference-based, human-based, and LLM-based approaches


\subsection{Evaluation Perspectives}


The primary objective of role-playing with language models is to facilitate interactions that are not only contextually relevant but also tailored to reflect specific personas or roles. Thus, evaluating such models involves multiple dimensions that collectively assess how well these agents perform their intended roles. These dimensions include: \textbf{conversation ability}, \textbf{role-persona consistency}, \textbf{role-behavior consistency} and \textbf{role-playing attractiveness}.

\subsubsection{Conversation Ability}
In theory, the concept of role-playing should extend beyond just dialogue, but up until now, all role-playing studies have been confined to role-playing in conversations. Therefore, general conversation ability plays a crucial role in the role-playing experience. Generally, the conversation ability required in role-playing is similar to that of open-domain dialogue models, but there can be exceptions. For instance, if a user asks the model to act as a character who doesn't speak, then this metric naturally becomes inapplicable.

\begin{itemize}[leftmargin=*]
    \item \textbf{Linguistic Quality}: This emerges as a crucial perspective, encompassing both \textit{fluency} and \textit{diversity} \cite{song-etal-2020-generate,Zheng_Zhang_Huang_Mao_2020}. Fluency assesses the grammatical correctness of responses, ensuring they are readable and free from obvious errors, while also considering the absence of repetitive structures and appropriate response length to effectively convey intended messages. Diversity, on the other hand, evaluates the richness of vocabulary used in responses. High lexical diversity signifies a model’s ability to vary its language use, which enriches dialogues, conveys nuanced information, and prevents conversations from becoming monotonous. Together, fluency and diversity in linguistic quality reflect the model's capacity to produce engaging and contextually appropriate responses that are both clear and varied, enhancing user interaction and satisfaction.
    \item \textbf{Coherence}: Coherence evaluates how relevant and logically consistent responses are with the ongoing conversation context \cite{chen2023large, song-etal-2021-bob}. It ensures that the dialogue flows naturally and that each response is contextually appropriate, maintaining continuity and relevance throughout the interaction. This involves the role-playing models' ability to track dialogue history.
\end{itemize}


\subsubsection{Role-Persona Consistency}

Another fundamental ability of role-playing models is to consistently represent a specific character or persona throughout interactions. Role persona mainly include two perspectives:

\begin{itemize}[leftmargin=*]
    \item \textbf{Attributes}: 
Typically, attributes  provide essential background information for language models to
play as a specific role, significantly influencing the character’s reactions
and interactions.
    Attributes define the persona of the agent, such as \textit{experiences, identities, interests, viewpoints, age, gender, achievements}, and \textit{ titles}. For instance, 
    a dialogue agent playing the role of a doctor would have attributes consistent with medical knowledge, a compassionate demeanor, and an authoritative title \cite{tu2024charactereval}.
    \item \textbf{Relations}: This involves the relationships between the speaker and others within the dialogue context. It considers how these relationships affect interaction dynamics, including familiarity, intimacy, animosity, or respect. For example, the relationship between Harry Potter and Ron is very different from the relationship between Harry and Voldemort \cite{chen2023large}.
\end{itemize}




\subsubsection{Role-Behavior Consistency}

Building on the solid base provided by persona consistency, role behavior consistency involves more dynamic elements that adapt during interactions. To keep language models' behavior consistent with a role requires a more sophisticated understanding of the context and ability to dynamically adjust, making the agent's actions feel more natural and appropriate.
It mainly includes:
\begin{itemize}[leftmargin=*]
    \item \textbf{Conversational Style}: The style of conversation should reflect the role's typical manner of speech \cite{zhou2023characterglm}. For example, a model playing a coach might use motivational and directive language, whereas one playing a friend might use a more casual and supportive style.

    \item \textbf{Personality}: The portrayal of traits of the model should be consistent with its role. Consistency in personality helps in forming a solid, believable character, which enhances user engagement and contributes to a seamless narrative flow within the system. For instance, an agent embodying Harry Potter exhibits a patient and encouraging attitude to Ginny, while might display competitiveness toward Draco Malfoy \cite{chen2023large}. 

    \item \textbf{Linguistic Features}: These include specific language use patterns, such as vocabulary, syntax, and register appropriate to the role. An elderly character might use more formal language and references from a different era, whereas a teenager might use slang and more relaxed grammar \cite{tu2024charactereval}.

\end{itemize}


\subsubsection{Role-Playing Attractiveness}

This is the most advanced level, where the role-playing models not only maintain persona and behavioral consistency but also enhances the interaction by being engaging, proactive, and empathetic:
\begin{itemize}[leftmargin=*]
    \item \textbf{Human Likeness}:
    The model should exhibit the naturalness of human interaction \cite{zhou2023characterglm}.  By interacting with what seem like real-world characters, users are more likely to engage deeply with the system. 
\item \textbf{Engagement}: Moreover, role-playing models are expected to actively keep the user interested and involved in the conversation, adapting its strategies based on user engagement levels \cite{chen2024roleinteract}.
\item \textbf{Proactivity}: This is another key feature of role-playing language models, which needs to actively initiate and drive conversations \cite{zhang2024unveiling}. It
involves the agent taking initiative within the conversation, suggesting topics and actions before the user expresses specific needs. This proactive response capability allows  role-playing models to anticipate user needs or interests and address them without explicit prompts from the user.

\item \textbf{Empathy}: It represents the peak of dialogue sophistication, where the agent recognizes emotional cues and responds in emotionally intelligent ways \cite{tu2024charactereval,zhang2024unveiling}.  Whether offering encouragement, advice, or sympathy, the response is tailored to the user's emotional state, thereby fostering a deeper emotional connection. This empathetic interaction is crucial for helping users address their emotional needs.
\end{itemize}

\subsubsection{Summary and Discussion}
Let's recall the essence of role-playing, which is to interact with users in a manner consistent with a specific role. Therefore, the most crucial metric for role-playing should be the degree of similarity to the role being portrayed. Hence, according to the authors' personal views and our experience in real-world applications, the more important dimensions among the above metrics should be Role-Persona consistency and Role-behavior consistency, as these two types of metrics truly measure the consistency of the LLM's behavior with the role. Furthermore, if there is no specific demand for the model, we recommend using fewer but more comprehensive metrics, which can both reduce the difficulty of annotation and facilitate the training of a more general reward model.

Indeed, in our practical engineering applications, we evaluate a general role-playing model solely based on a single metric: model-role similarity, to measure the alignment degree between the model and the role in any scenarios. Furthermore, when considering launching this model in a specific region, we use safeness to evaluate whether the model complies with the laws and regulations of that region. When the model is used to play a game NPC, we evaluate whether the model's answers contain knowledge hallucinations that are inconsistent with the game background.

\subsection{Evaluation Approaches}

To evaluate role-playing language models on the aforementioned dimensions, existing methods can be categorized into three main types: Reference-based, human-based, and LLM-based evaluation. The last type of evaluation method has emerged with the introduction of ChatGPT, and its principle is to use LLM to simulate human evaluation.

 \subsubsection{Reference-based Evaluation} 

 The reference output (ground truth) in the test set represents what users expect to see from the model output, essentially reflecting the most authentic user needs. Therefore, based on the assumption that `the more similar the model's output is to the reference, the better the model performs', researchers have begun to use some reference-based metrics to evaluate the consistency between the model output and the test set.
 This is often associated with metrics like Perplexity (PPL) \cite{zhang-etal-2018-personalizing}, BLUE~\cite{papineni-etal-2002-bleu}, and ROUGE~\cite{lin-2004-rouge}. PPL is used to estimate the likelihood of the reference output on the evaluated model, while BLEU and ROUGE are used to measure the similarity between the generated output and the reference. The key difference between the former and the latter two lies in the fact that PPL measures the alignment at the distribution level between the model and the reference, independent of the decoding algorithm. On the other hand, BLEU and ROUGE require the model to first decode a text output, and then evaluate the similarity between the two pieces of text.

 
 Although we already have a variety of commonly used reference-based metrics, they still fall short in evaluating how well the responses align with personas or characters. To address this, several methods are explored the multi-choice answer format evaluation \cite{salemi2024lamp, shen2024roleeval}, where LLM need to choose the most persona-consistent response from given options rather than open-end generation, measuring accuracy directly. 

An alternative approach emphasizes the use of relative automatic metrics, such as Delta PPL, to gauge the model's role-playing capability using triplets like \((x, y_{\text{win}}, y_{\text{lose}})\). Specifically, it considers the difference in Perplexity (PPL)~\cite{song-etal-2021-bob}  between positive and negative instances in the test set as a measure of the model's role-playing proficiency. 

To the best of our knowledge and experience, Delta PPL is the sole reference-based model capable of objectively and accurately assessing an LLM's role-playing ability. However, this metric is heavily dependent on a meticulously curated test set comprising data in triplet form.

  \subsubsection{Human-based Evaluation} This is critically important in assessing role-playing models because automatic metrics often fail to capture the nuanced aspects of responses that are essential for realistic and engaging character simulation. Therefore, human-based evaluation is considered one of the most effective methods for thoroughly assessing performance. 

  To the best of our knowledge, within the realm of industrial applications, human evaluation stands as the most crucial, if not the only reliable, method of evaluation. This is primarily because only humans can authentically replicate the experiences of other human users, thereby offering insights into the actual user experience.

  Theoretically, all the perspectives introduced in the previous section can be evaluated by human annotators, but generally, due to cost and the limitations of annotators' capabilities, it is recommended to integrate multiple dimensions into a single dimension for evaluation. In terms of implementation, human-based evaluations typically consist of two approaches:
  
  (1) \textbf{Scoring}: In this method, each relevant dimension is assigned a specific scoring threshold, and annotators rate the model’s responses based on these criteria \cite{zhou2023characterglm, tu2024charactereval}. Common manual scoring metrics for evaluating role-playing models encompass several key aspects: \textit{Consistency}, which ensures that responses align with the attributes and behaviors specified in the character's profile; \textit{Human-likeness,} assessing how well the responses mimic human characteristics and natural communication; and \textit{Engagement}, measuring the ability of the response to capture attention or arouse curiosity. Additionally, metrics like \textit{coherence} and \textit{fluency} are typically included to evaluate how well responses fit into the overall dialogue context and their linguistic smoothness, respectively. To aid in this evaluation process, it is also common practice to provide human annotators with examples for each criterion, helping them to better understand and apply the evaluation standards consistently and effectively. This structured approach ensures a comprehensive assessment of the role-playing model's performance across multiple dimensions of dialogue quality. But it is also complex
  due to the need for detailed scoring samples for each rating. It also adds to the complexity of the evaluation process and might not facilitate easy comparisons across multiple models. 
  
  (2) \textbf{Ranking}: Another alternative approach is to present responses from multiple models to annotators who then rank these in order of quality or categorize them into win/tie/lose outcomes \cite{chen2023large,chen2024compress}. A typical example of this method in use is with instructGPT \cite{DBLP:conf/nips/Ouyang0JAWMZASR22}, where human labelers rank the model’s outputs from best to worst based on quality. Similarly, \citet{chen2023large}  employ annotators to rank various models in terms of their role-playing capabilities, specifically evaluating them based on their relevance with scenes, characters attributes, and relations. This ranking approach allows for a direct comparison of models, giving a clearer picture of their respective strengths and weaknesses in simulating realistic and contextually appropriate role-playing interactions.
  Ranking is more efficient and can compare multiple models simultaneously, but it generally provides less granularity than scoring because it aggregates the quality into broader categories.

Despite our earlier emphasis that ``human evaluation is the only reliable method of evaluation'', its practical application is often beset with challenges due to numerous inherent shortcomings. The most critical issue stems from the subjectivity of human annotators. This subjectivity can result in: 1) Significant bias, particularly when the annotation process lacks rigor. A typical scenario we frequently encounter involves model developers or paper authors inviting their colleagues or friends to act as annotators. In such cases, these individuals are naturally inclined to award higher scores to `the proposed method.' 2) The inability to directly reuse the human-evaluation results from one paper in another, thereby compelling each paper to invest substantial time and resources in conducting human evaluations for all baselines.

\subsubsection{LLM-based Evaluation}

In light of the aforementioned issues with human evaluation, some researchers have started experimenting with LLMs as annotators for role-playing tasks~ \cite{chen2023large, yu2024neeko}. Undoubtedly, employing LLMs as annotators can mitigate the issues of bias and cost to some extent. However, this approach gives rise to a new question: Are LLMs sufficiently competent as annotators for role-playing tasks?~\cite{wang2023chatgptgoodnlgevaluator}

In practice, LLM-based evaluation involves detailed instructions  that include evaluation dimensions and thresholds, which generally align with the criteria used in human assessments \cite{chen2023large,chen2024compress}. To guide the LLM towards more accurate evaluations, several related scoring examples are often provided and formulated in a predefined format that  facilitates straightforward aggregation and analysis of the results.
In general, LLM-based evaluations reduce the need for extensive human annotator training and coordination, providing a scalable option for accessing models effectiveness. Despite the speed and reduced logistical overhead, LLM-based evaluations often face challenges in achieving consistency with human judgments. When LLMs are tasked with evaluating a role they are not familiar with, the accuracy of the evaluation may be compromised. There are also several general shortcomings, such as LLMs' sensitivity to order when scoring, often giving a higher rank to responses placed earlier. Furthermore, LLMs tend to assign higher ranks to lengthier responses. 
Lastly, LLMs typically struggle to accurately evaluate models that possess superior role-playing capabilities than their own. For example, a reward model based on ChatGPT would not be able to accurately assess the capabilities of a role-playing model based on GPT-4. In such scenarios, if the goal is to train a role-playing model that surpasses the state-of-the-art performance, the use of LLM-based evaluation becomes unsuitable.

In some recent studies, we have observed a concerning trend: \textbf{researchers are increasingly relying on LLM-based evaluation without adequately verifying the effectiveness of this metric for their specific tasks}. In fact, we have meticulously replicated some of the methods used for evaluation in certain papers, and regrettably found that the effectiveness of these methods is actually far behind that of human experts\footnote{To prevent any potential negative implications for the authors of these papers, we refrain from citing them here, but they are all included in the references of this paper}. Such low-quality evaluations can trigger a series of chain reactions, such as the follow-up work of these studies continuing to use these evaluation methods. Consequently, we make this appeal: authors must provide ample evidence within their papers to substantiate that their LLM-based evaluation method is an appropriate evaluator for their specific scenarios. For example, they could select a small sampled set and illustrate the consistency between well-instructed human annotators and LLM evaluators on these samples. As contributors of knowledge to the community, it is essential for the authors to ensure the accuracy and validity of this knowledge.

\subsubsection{Evaluating RPLAs}  Evaluating RPLAs, especially in scenarios involving the collaboration of multiple agents such as MetaAgents \cite{li2023metaagents} or GenerativeAgents \cite{park2023generative}, presents unique challenges. In these settings, the effectiveness of these models in role-playing is assessed based on their ability to achieve predefined goals or complete specific tasks. 
The evaluation process typically focuses on whether the agents can successfully fulfill the roles and objectives they are assigned, which is crucial in settings where multiple agents must work together coherently. 
Metrics such as accuracy are commonly used to quantitatively measure how well these agents meet their goals.
Additionally, qualitative evaluations might involve human judges assessing the appropriateness and effectiveness of specific actions taken by the agents within the scenario. For instance, in the simulation of human behaviors, the evaluation of RPLAs  often involves comparing their actions to those of human-annotated standards to determine their accuracy and effectiveness. 

\subsubsection{Summary and Discussion} Evaluating role-playing language models effectively often involves a mix of methods, each with its own strengths and weaknesses. \textbf{Reference-based metrics} are efficient and objective, providing quick, quantifiable results ideal for preliminary assessments, though they lack depth and context sensitivity, failing to capture nuances like persona consistency. \textbf{Human-based evaluations} offer deep insights into nuances and subtleties in dialogues, including character alignment and user engagement, but are costly and less scalable, with potential for subjective variability between evaluators. \textbf{LLM-based evaluations}, leveraging the capabilities of large language models, offer scalability and speed and can mimic some aspects of human judgment, yet they may not always align with human evaluations and depend heavily on the used LLMs. 

Finally, we want to emphasize a principle that is the most fundamental for role-playing evaluation: regardless of the evaluation method adopted, one must meticulously verify that the method is capable of evaluating cases in their specific scenarios.  To illustrate, consider these simple examples: when carrying out human \& LLM-based evaluations, we must ensure that the annotators \& LLMs possess sufficient background knowledge of the roles to be evaluated; otherwise, they lack the necessary capability to evaluate this task. On the other hand, when using the $\Delta$ PPL metric for evaluation, we must also ensure that the quality of the positive response is indeed higher than that of the negative response.


%% file: sections/6Challengs_Future.tex
\section{Challenges and Future Directions}
\label{challenge}

Building role-playing language models face a myriad of challenges that impact their development and effectiveness in delivering complex, interactive narratives. While these models or agents have shown promising capabilities, significant hurdles remain in metrics development, evaluation accuracy, and system adaptability. This section explores these challenges in detail, identifying key areas where current systems fall short and proposing directions for future research. Please note that we have already introduced some challenges earlier in the text. In this section, we will bring all these elements together for a comprehensive summary.

 \paragraph{More Reference-based Metrics for evaluating Role-Playing.} Current reference-based evaluation metrics for role-playing primarily focus on linguistic accuracy and coherence, which are insufficient for assessing role-playing capabilities that require character consistency and narrative engagement \cite{chen2023large}. To the best of our knowledge and experience, 
$\Delta$ PPL~\cite{song-etal-2021-bob} is the sole reference-based model capable of objectively and accurately assessing an LLM’s role-playing ability. However, methods based on PPL are only capable of evaluating effects at the distribution level, and they fall short when it comes to assessing the quality of the final generated results, which are derived from decoding the output distribution. To our knowledge, as of now, there is no existing method that can directly evaluate the consistency between the assigned role and the generated output.
 

 \paragraph{Sensitivity in LLM-Based Evaluation.} 
 LLM-based evaluations often face challenges in achieving consistency with human judgments. When LLMs are tasked with evaluating a role they are not familiar with, the accuracy of the evaluation may be compromised. There are also several general shortcomings, such as LLMs' sensitivity to order when scoring, often giving a higher rank to responses placed earlier. Furthermore, LLMs tend to assign higher ranks to lengthier responses. 
Lastly, LLMs typically struggle to accurately evaluate models that possess superior role-playing capabilities than their own. For example, a reward model based on ChatGPT would not be able to accurately assess the capabilities of a role-playing model based on GPT-4. In such scenarios, if the goal is to train a role-playing model that surpasses the state-of-the-art performance, the use of LLM-based evaluation becomes unsuitable.

 \paragraph{Imbalance, Bias and Cost in Human-based Evaluation.} 
 While human evaluation is invaluable for capturing the nuance and complexity of role-play interactions, it is resource-intensive and difficult to standardize. First, annotators could exhibit bias in their evaluation, such as position biases, the
preference for verbose and complex responses \cite{pandey2022modeling, santurkar2023whose}.
On the other hand, training evaluators to consistently assess role-playing performance is challenging, particularly when dealing with subjective interpretations of character and narrative \cite{chen2023large}. For instance, validating the models' performances of playing Harry Potter requires that human evaluators be familiar with the magic worlds and characters relationships with Harry in different storylines. 
 Moreover, balancing and calibrating human judgments to ensure reliability across diverse scenarios adds another layer of complexity \cite{ethayarajh2022authenticity}. Developing more structured evaluation frameworks and training protocols could help mitigate these issues \cite{prassl2017legal,clark2021all}.


 \paragraph{Lack of deeper Role-specific Alignment Approaches. } Aligning language models with specific roles, particularly using LLMs, presents significant challenges. Current approaches primarily integrate persona and interaction context directly into input prompts or instructions \cite{tu2023characterchat,zhou2023characterglm}. However, these methods often lack depth in modeling the complexities of character roles, such as relationships between characters, their psychological states, or evolving dynamics throughout an interaction \cite{chen2023large}. This surface-level adaptation does not allow models to learn and adapt to the nuanced interactions that define characters within narratives, limiting the models' ability to deliver immersive and contextually rich interactions.  \citet{ahn2024timechara} prove that understanding and replicating how a character’s emotional response might evolve in reaction to another character's actions or environmental changes are areas that are seldom explored in depth. Future research needs to focus on developing techniques that enable role-playing language models to internalize and dynamically represent interpersonal relationships and psychological states of characters. This could involve sophisticated training regimes that incorporate dynamic character-based scenarios, psychological profiling, and relationship mapping. Such advancements would enhance the realism and engagement in role-playing, making them more effective for applications in gaming, training simulations, and interactive storytelling, where accurate and dynamic role portrayal is crucial for enhancing user experience.


 \paragraph{Ensure the safety in Role-Playing.} Ensuring safety in role-playing language models involves addressing several key areas: privacy \cite{chen2024compress}, toxicity \cite{wen2023unveiling,gehman-etal-2020-realtoxicityprompts,deshpande2023toxicity}, and biased or discriminatory behavior \cite{nangia2020crowspairs,rutinowski2023selfperception,shaikh2023second}.  Privacy concerns are paramount, especially on platforms where users interact extensively with agents. These agents often process and store sensitive personal information derived from interactions to enhance the role-playing experience. Ensuring that this information is not misused or inadvertently leaked is critical. Techniques such as data anonymization \cite{eurojust2019cybercrime}, secure data storage, and strict access controls are necessary to protect user privacy. Additionally, implementing protocols that ensure data is only used to enhance the interaction, without being stored longer than necessary, can help maintain user trust.

 Toxic role-related content  often stems from the data used during the pre-training and fine-tuning phases, which may include biased or harmful language from the internet or other sources \cite{rutinowski2023selfperception}. This is particularly problematic in role-playing scenarios, where agents are expected to adapt to diverse personas that may inadvertently include or trigger negative content. To counter this, it's essential to employ robust content moderation systems and retrain models using curated, non-toxic datasets. Continuous monitoring and updating of the model’s responses using feedback loops can also help minimize the emergence of undesirable content.

 Moreover,  role-playing language models can exhibit biased or discriminatory behavior if their training data contains implicit cultural or societal biases \cite{brown2023scalable}. \citet{ma2024visualroleplayuniversaljailbreakattack} further conduct Jailbreak Attack on MultiModal LLMs via Role-playing Image Character. 
 Such phenomenon is especially concerning in role-playing, where such behavior can significantly detract from the user experience and perpetuate harmful stereotypes. To address this, diversity and inclusion must be integral to the training process. Techniques like fairness audits, bias testing, and incorporating diverse datasets in training can help reduce these issues. Furthermore, regular updates and checks on the model's outputs against established fairness standards can ensure more balanced interactions.

 \paragraph{Hallucination in Role-Playing.}  Hallucination in LLMs, particularly pronounced in knowledge-intensive tasks, poses significant challenges for LLMs to generate knowledge-consistent responses \cite{ji2023survey,xu2024hallucination}. This issue, known as \textbf{character hallucination} \cite{shao2023character,ahn2024timechara,sadeq2024mitigating}, occurs when the language models generate responses that are inconsistent with the defined profiles or historical context of a character. For example, a model designed to role-play Queen Elizabeth I using modern slang or discussing contemporary issues would be a clear instance of character hallucination due to its anachronistic and contextually inaccurate content. A more complex aspect, point-in-time character hallucination \cite{chen2023large}, involves maintaining narrative consistency over time, such as ensuring a character's responses evolve correctly according to their development in the storyline.

 To mitigate these issues, effective strategies include fine-tuning within character-related domain knowledge \cite{tian2023chatplug}, where models are specifically adjusted to reflect accurate character and narrative details, and RAG, which supplements the model’s responses with real-time, role-specific information. These approaches enhance the model’s ability to generate temporally and contextually appropriate responses, thereby reducing inaccuracies and improving the overall reliability and immersion of language models in role-playing scenarios. Addressing such issues is crucial for developing sophisticated role-playing agents that accurately embody characters and maintain narrative fidelity throughout interactions.

 \paragraph{Memory Usage in RPLAs.} Memory module in RPLAs is essential for providing contextual continuity, role-playing interactions, and deep narrative engagement. However, the implementation faces  significant challenges, particularly in\textit{ managing large volumes of data} \cite{wang2023recursively}, \textit{ensuring the accuracy of compressed memories} \cite{chen2024compress}, \textit{efficiently updating memory databases} \cite{bae-etal-2022-keep}, and \textit{achieving precise memory retrieval} \cite{DBLP:conf/acl/XuGWNW0W22}. One primary challenge is handling the extensive input context length, which can overwhelm the model if all past interactions are directly inputted as memory. To manage this, RPLAs often use compressed versions of past interactions \cite{wang2023recursively, park2023generative}, which must be carefully crafted to ensure no critical information is lost. Furthermore, the operation of memory databases requires sophisticated algorithms for timely updates and appropriate forgetting of irrelevant data to maintain efficiency and relevance. Retrieval accuracy also poses a major challenge, especially in long-term real-world scenarios where the precision of fetched memories significantly impacts the quality of interactions. To address these issues, researchers have proposed various innovative solutions. \citet{zhong2024memorybank} suggest using a memory bank approach where memory updating is governed by a forgetting curve, which helps in phasing out outdated or less useful memories over time. Others advocate for multiple compressions of memory to reduce dependency on retrieval models \cite{chen2024compress}, thereby minimizing the risks of retrieving inaccurate memories.  
 
 Despite these efforts, memory management in RPLAs remains a complex, ongoing issue that requires continued research and development. Future advancements will need to focus on refining these strategies, ensuring that RPLAs can effectively recall and utilize memories to enhance user interaction without compromising system performance.

\paragraph{Multi-Modal Integration in RPLAs.}

While the majority of RPLAs traditionally operate with text-based inputs and outputs, incorporating multi-modal interactions can significantly enhance the depth and realism of role-playing. 
Integrating image-text pairs allows models to learn more detailed character traits, as demonstrated in \cite{ahn2023mpchat, dai2024mmrolecomprehensiveframeworkdeveloping}. Similarly, \citet{zhao2023narrativeplay} leverages models like stable diffusion to generate enriched environment in narrative role-playing where textual information is insufficient, adding depth to the storytelling.
The integration of multi-modal interactions represents a promising frontier for enhancing the capabilities and effectiveness of RPLAs. As progresses, the potential to seamlessly blend visual, auditory, and textual data will likely expand, opening up new avenues for creating more immersive and realistic role-playing experiences.

 \paragraph{Make RPLAs be lifelong learner.} Lifelong learning represents a pivotal long-term goal for RPLAs, aiming to enable these systems to continuously adapt and evolve in response to user interactions and environmental changes \cite{park2023generative, wang2023voyager}. This capability is crucial for maintaining the relevance and effectiveness of RPLAs over extended periods. 

 However, the potential for lifelong learning introduces significant challenges, particularly in maintaining alignment and safety. As RPLAs evolve, they may develop capabilities that lead to unintended or harmful outcomes, such as reward hacking or misaligned goals, which could manipulate monitoring systems or pursue unethical strategies \cite{shevlane2022structured,shevlane2023model}. Ensuring that RPLAs remain aligned with ethical standards and their intended roles requires continuous oversight and adaptive mechanisms to correct and guide their learning trajectories \cite{brown2023scalable,cohen2022advanced}. Developing robust frameworks to manage lifelong learning—incorporating advanced monitoring and adjustment techniques—will be crucial for the successful integration of RPLAs into various domains, from entertainment to customer service, ensuring they remain beneficial and safe as they evolve.

\subsection{Summary and Discussion} In general, the advancement of role-playing language models faces numerous challenges, including the development of specific evaluation metrics, efficient memory management, ensuring role alignment, maintaining safety, and facilitating lifelong learning. Each of these areas presents unique obstacles that impact the effectiveness and safety in dynamic environments. Addressing these issues requires an integrated approach that combines technological innovation with strict adherence to ethical standards, ensuring the model not only perform their roles with enhanced accuracy and realism but also operate safely and responsibly. Future progress in this field will depend on our ability to balance these complex factors, paving the way for role-playing language models to transform interactive storytelling and digital interactions across various applications.
 

%% file: sections/7Conclusion.tex
\section{Conclusion}
The development of role-playing with language models represents a significant evolution in the field of artificial intelligence. From their inception, where models primarily focused on maintaining simple persona consistencies, to the present day, where advanced LLMs facilitate complex and nuanced character-driven interactions, the progress has been substantial. This survey highlights key advancements in data sourcing, foundation models, alignment, system architecture, and evaluation, demonstrating that current models not only maintain character consistency but also exhibit dynamic behavior and attractiveness. Despite the progress, challenges such as dynamic persona management, behavioral alignment, ethical considerations, and the development of comprehensive evaluation metrics persist. Addressing these challenges will be crucial for future advancements, aiming to create more immersive and realistic role-playing applications with LLMs.